\documentclass[acmlarge]{acmart}

\usepackage{multirow}
\usepackage{graphicx}
\usepackage{caption}
\usepackage{subcaption}
\usepackage{dblfloatfix}
\usepackage{xcolor}
\usepackage{booktabs}
\usepackage{wrapfig}
\usepackage{soul}
\usepackage{enumerate}
\usepackage{enumitem}
\usepackage{makecell}
\usepackage[table]{xcolor}

\usepackage{tikz}
\usepackage{pgfplots}
\pgfplotsset{compat=1.18}

\acmJournal{IMWUT}
\acmVolume{0}
\acmNumber{0}
\acmArticle{0}
\acmMonth{0}

\begin{document}

\title{A Comparative Study of Traditional Machine Learning, Deep Learning, and Large Language Models for Mental Health Forecasting using Smartphone Sensing Data}

\author{Kaidong Feng}
\affiliation{%
  \institution{Singapore University of Technology and Design}
  \city{Singapore}
  \country{Singapore}
}
\email{kaidong3762@gmail.com}

\author{Zhu Sun}
\affiliation{%
  \institution{Singapore University of Technology and Design}
  \city{Singapore}
  \country{Singapore}
}
\email{zhu_sun@sutd.edu.sg}

\author{Roy Ka-Wei LEE}
\affiliation{%
  \institution{Singapore University of Technology and Design}
  \city{Singapore}
  \country{Singapore}
}
\email{roy_lee@sutd.edu.sg}

\author{Xun Jiang}
\affiliation{%
  \institution{Tianqiao and Chrissy Chen Institute}
  \city{Singapore}
  \country{Singapore}
}
\affiliation{%
  \institution{Theta Health Inc}
  \city{Singapore}
  \country{Singapore}
}
\email{jiangxun@shanda.com}

\author{Yin-Leng Theng}
\affiliation{%
  \institution{Nanyang Technological University}
  \city{Singapore}
  \country{Singapore}
}
\email{tyltheng@ntu.edu.sg}

\author{Yi Ding}
\affiliation{%
  \institution{Purdue University}
  \city{West Lafayette}
  \country{USA}
}
\email{yiding@purdue.edu}

\begin{abstract}
	
Smartphone sensing offers an unobtrusive and scalable way to track daily behaviors linked to mental health, capturing changes in sleep, mobility, and phone use that often precede symptoms of stress, anxiety, or depression. While most prior studies focus on detection that responds to existing conditions, forecasting mental health enables proactive support through Just-in-Time Adaptive Interventions. In this paper, we present the first comprehensive benchmarking study comparing traditional machine learning (ML), deep learning (DL), and large language model (LLM) approaches for mental health forecasting using the College Experience Sensing (CES) dataset, the most extensive longitudinal dataset of college student mental health to date. We systematically evaluate models across temporal windows, feature granularities, personalization strategies, and class imbalance handling. Our results show that DL models, particularly Transformer (Macro-F1 = 0.58), achieve the best overall performance, while LLMs show strength in contextual reasoning but weaker temporal modeling. Personalization substantially improves forecasts of severe mental health states. By revealing how different modeling approaches interpret phone sensing behavioral data over time, this work lays the groundwork for next-generation, adaptive, and human-centered mental health technologies that can advance both research and real-world well-being.
	
\end{abstract}

%


\maketitle

\section{Introduction}\label{sec:intro}

Ubiquitous computing technologies are increasingly embedded into our daily lives~\cite{bardram2022sensing}. Smartphone and wearable devices continuously and passively capture fine-grained traces of daily behavior~\cite{abdullah2014towards,nelson2019accuracy,boe2019automating}. Unlike traditional self-report surveys or clinical assessments that rely on infrequent and subjective inputs, passive sensing provides unobtrusive, continuous, and real-world measurements of a person’s daily routines and social interactions~\cite{coravos2019developing,dunn2021wearable,master2022association,nepal2024social}. This continuous data stream is particularly valuable for mental health, where symptoms such as stress, anxiety, and depression often manifest subtly through changes in behavior, sleep, movement, or communication before individuals are consciously aware of them.

Over the past decade, a growing body of research has shown that longitudinal passive sensing can effectively support mental health monitoring~\cite{wang2014studentlife} and detection~\cite{xu2022globem,roychowdhury2025predicting,adler2021identifying,barnett2018relapse,wang2016crosscheck,adler2024measuring,meegahapola2023generalization,meegahapola2024m3bat,zhang2024reproducible}. For example, fluctuations in mobility and communication have been linked to social withdrawal~\cite{stamatis2023specific}; changes in phone use or sleep timing have been associated with anxiety or depressive episodes~\cite{mohd2024relationship}. However, the behavioral signals captured by smartphones and wearables are multimodal, noisy, and highly individualized~\cite{vaid2021ubiquitous}, requiring computational methods capable of extracting stable and interpretable patterns over time.

To address this complexity, researchers have increasingly leveraged machine learning (ML) techniques for mental health detection. \emph{Detection} research builds ML models that identify individuals who are currently experiencing a mental health condition or related symptoms based on sensing data~\cite{farhan2016behavior,wang2018tracking}. 
Detection is valuable because it relieves users from the burden to self-report and can automatically infer their current mental state. Traditional ML models (e.g., random forests, support vector machine)~\cite{chikersal2021detecting, zhu2023stress,opoku2021predicting,lewis2023mixed,shen2025passive} have been used to identify correlations between behavioral features and mental states. More recent deep learning (DL) approaches, such as recurrent and transformer-based models, can learn nonlinear temporal dependencies across multimodal sensing streams, enabling more accurate prediction of mental health trajectories~\cite{fukazawa2020smartphone,yang2022survey,vos2023generalizable,umematsu2019daytime}.  The latest frontier involves large language models (LLMs), which offer a flexible and scalable way to model heterogeneous data sources and reason about behavioral context~\cite{chen2025unveiling, xu2024mental, kim2024healthllm}.  
Despite these progresses, detection is reactive as it identifies problems only after they have occurred, rather than anticipating them before they emerge.

In this paper, we conduct a comparative study of traditional ML, DL, and LLM methods for mental health forecasting problem. Unlike detection, which reacts to existing conditions, \textbf{forecasting} aims to predict future mental health changes from passive sensing data before symptoms become apparent. We focus on forecasting because it is more practical and challenging, enabling Just-in-Time Adaptive Interventions (JITAI)~\cite{nahum2017just,klasnja2019efficacy} that can act proactively to prevent crises. Our study builds on the College Experience Sensing (CES) dataset~\cite{nepal2024capturing}, the most extensive longitudinal passive sensing dataset of college student mental health, released in October 2024. CES captures rich behavioral and contextual signals over multiple years, enabling us to evaluate model performance under realistic, long-term conditions. To the best of our knowledge, this is the first study to compare traditional ML, DL, and LLM approaches for mental health forecasting on the CES dataset. The previous study, I-HOPE~\cite{roychowdhury2025predicting}, addressed only detection on the CES dataset, leaving the forecasting problem unexplored. 

To guide this investigation, we first formulate six research questions (RQs):
\begin{itemize}[leftmargin=0.5cm]
    \item RQ1: How do traditional ML, DL, and LLM-based approaches compare in forecasting mental health states?
    \item RQ2: How do different LLM adaptation strategies (i.e., Zero-shot~\cite{xu2024mental}, In-Context Learning (ICL)~\cite{kim2024healthllm}, and Parameter Efficient Fine-Tuning (PEFT)~\cite{han2024parameter}) influence forecasting accuracy?
    \item RQ3: How robust are the models in early forecasting when the available behavioral history is limited?
    \item RQ4: How do varying feature representation strategies across dimensional and temporal granularity affect model performance?
    \item RQ5: To what extent does personalization (user-specific modeling) improve forecasting effectiveness?
    \item RQ6: Does the dataset exhibit severe class imbalance? If so, how do different loss functions such as Weighted Cross-Entropy~\cite{aurelio2019learning} and Focal Loss~\cite{lin2017focal} mitigate this issue during training? 
\end{itemize}

Together, these research questions aim to develop a holistic understanding of what works, when, and why in forecasting mental health using phone sensing behavioral data. To address them, we systematically design a comprehensive set of experiments. Specifically, we formulate the mental health forecasting task as a multi-class classification problem using the CES dataset, where the objective is to predict a participant's future PHQ-4~\cite{kroenke2009ultra} severity level (Normal, Mild, Moderate, or Severe) at the end of the second week, based on their behavioral data from the previous week.
We carefully select representative methods from each modeling approach. For traditional ML models, we employ Logistic Regression~\cite{hosmer2013applied}, Support Vector Machine~\cite{brereton2010support}, Random Forest~\cite{breiman2001random}, Decision Tree~\cite{song2015decision}, XGBoost~\cite{chen2016xgboost}, and LightGBM~\cite{ke2017lightgbm} as interpretable, feature-based baselines. For DL, we implement Multi-Layer Perceptron (I-HOPE~\cite{roychowdhury2025predicting}), Temporal Convolutional Networks ~\cite{bai2018empirical}, Long Short-Term Memory with Attention~\cite{yu2019review}, and Transformer~\cite{vaswani2017attention} to capture temporal behavioral dynamics. For LLMs, we evaluate Qwen3 (4B, 8B, and 14B) and GPT-4.1 under different adaptation strategies, including Zero-shot~\cite{xu2024mental}, In-Context Learning (ICL)~\cite{kim2024healthllm}, and Parameter-Efficient Fine-Tuning (PEFT)~\cite{han2024parameter}.
To further explore model behavior, we conduct an incremental temporal evaluation simulating expanding observation windows to assess early prediction capability. We also systematically examine the effects of feature representation granularity (35-Dimension vs. 5-Dimension; Daily vs. Weekly Aggregation), personalization mechanisms (user embeddings or identifiers), and class imbalance mitigation techniques (Weighted Cross-Entropy~\cite{aurelio2019learning} vs. Focal Loss~\cite{lin2017focal}).

Several key findings emerge from our comprehensive experiments.
\textbf{(1)} DL models outperform all other approaches, achieving the highest overall performance (Macro-F1 = 0.5808 by the Transformer), whereas traditional ML models remain competitive for the majority class (i.e., Normal) but struggle with rare or overlapping mental health states such as Mild and Moderate.
\textbf{(2)} LLMs exhibit moderate forecasting accuracy, generally lower than that of DL and traditional ML models, due to their limited capacity to model numerical and temporal dependencies. However, they excel at semantic reasoning, where Few-shot Learning strategies consistently outperform both Zero-shot and PEFT approaches.
\textbf{(3)} Early forecasting is feasible: DL models and LLMs demonstrate stronger early prediction capability compared to traditional ML methods, suggesting their potential for timely mental health intervention.
\textbf{(4)} Feature granularity plays a critical role in forecasting outcomes, with most methods benefiting from more compact and aggregated feature representations.
\textbf{(5)} Personalization markedly enhances forecasting performance, particularly for severe cases. DL models augmented with user embeddings achieve Macro-F1 improvements greater than 0.3, underscoring the importance of personalization for individualized mental health prediction.
\textbf{(6)} Across all models, the majority class (i.e., Normal) consistently achieves substantially higher accuracy than the minority classes (e.g., Severe), and Focal Loss outperforms Weighted Loss for high-capacity DL models by emphasizing minority and difficult samples, thereby mitigating class imbalance more effectively.

In summary, this work makes three key contributions
\begin{itemize}[leftmargin=*]
    \item We present the first systematic benchmarking study that compares traditional ML, DL, and LLM-based approaches for mental health forecasting using the CES dataset - the most extensive and temporally rich longitudinal dataset on college student mental health to date.
    \item We conduct a unified, multi-perspective evaluation framework that jointly examines temporal sensitivity, feature granularity, personalization, and class imbalance, providing an integrated understanding of how different modeling approaches behave under real-world, noisy sensing conditions.
    \item Our study yields several insightful findings and identifies potential challenges that offer practical design implications for next-generation, human-centered mental health forecasting systems. These insights can inform the development of models that are not only accurate and data-efficient but also more interpretable, adaptive, and ethically deployable in real-world contexts. 
\end{itemize}

\section{Related Work}

We first review prior studies that apply LLMs to text-based mental health detection, and then summarize research on machine/deep learning methods for mental health detection and forecasting using mobile sensing data.

\subsection{LLMs for Mental Health}

Recent research has begun exploring the potential of LLMs for analyzing emotions and mental states. Some studies showed that LLMs can perform sentiment analysis and emotion reasoning across diverse textual contexts~\cite{kocon2023chatgpt,qin2023chatgpt,zhong2023can,yang2023towards}.
Building on this line of work, \citet{amin2023affective} and \citet{lamichhane2023evaluation} evaluated ChatGPT (GPT-3.5) on several mental health classification tasks such as stress, depression, and suicide detection. Their results revealed that ChatGPT shows initial promise for mental health applications but still lags behind specialized models by 5–10\% in accuracy and F1-score. \citet{yang2023evaluations} investigated ChatGPT's reasoning ability on tasks requiring causal or contextual inference, such as identifying potential stressors. More recently, Mental-LLaMA~\cite{yang2024mentallama} was introduced, which is a family of LLaMA-based models fine-tuned on mental health datasets for a range of related tasks. In parallel, Health-LLM~\cite{kim2024healthllm} and Mental-LLM~\cite{xu2024mental} were introduced that benchmark multiple LLMs and fine-tuning methods (zero-shot prompting, few-shot prompting, and instruction fine-tuning) across diverse mental health datasets. However, these studies have focused on text-based detection tasks, leaving the potential of LLMs for modeling numerical sensing data for forecasting tasks largely unexplored.

\subsection{Machine/Deep Learning for Mental Health Detection Using Mobile Sensing Data}

Mental health conditions such as depression, stress, and emotional dysregulation affect hundreds of millions of people worldwide and remain a growing public health concern. For example, depression affects over 216 million people globally and approximately 7.2\% of U.S. adults annually, with prevalence rates rising during and after the COVID-19 pandemic~\cite{huckins2020mental,nepal2022covid}. Traditional screening methods rely on self-reports or clinical assessments, which are often infrequent, subjective, and resource-intensive. In contrast, the increasing ubiquity of smartphones and wearable devices has enabled continuous, passive, and unobtrusive monitoring of human behavior in natural environments~\cite{wang2016crosscheck,aung2016leveraging,canzian2015trajectories}. These devices record diverse behavioral and physiological signals, such as mobility, sleep, phone use, communication, and heart rate, that can serve as digital biomarkers of mental health.

Methodologically, research in mobile sensing for mental health has evolved through several key stages.
Early work primarily conducted statistical correlation analyses to establish relationships between behavioral signals and mental health outcomes. For instance, \citet{saeb2015mobile} linked depressive symptom with mobility and phone usage patterns, while \citet{ben2015next} found correlations between depression changes and sleep duration, speech activity, and mobility. Similar studies extended such analyses to stress and emotion sensing, identifying significant relationships between daily routines, social activity, and affective states. These early findings validated the feasibility of using passive sensing data to capture behavioral cues associated with mental well-being.

As datasets became larger and more longitudinal, researchers began developing machine learning (ML) models to move beyond correlation and enable automatic detection of mental health status. Traditional ML approaches, such as support vector machines, random forests, and logistic regression, were used to classify depressive or stress-related states from handcrafted behavioral features~\cite{farhan2016behavior,wang2018tracking}. For example, \citet{farhan2016behavior} detected biweekly depression using location features from 79 students. \citet{wang2018tracking} combined multimodal features from smartphones and wearables to achieve high recall in depression detection among college populations. Later, \citet{xu2019leveraging} applied automated feature extraction via association rule mining and introduced personalized collaborative-filtering models that improved detection performance.

More recent work has adopted deep learning (DL) architectures, including recurrent neural networks and transformers, to capture nonlinear and long-term temporal dependencies across multimodal sensing streams. These models have been applied to various emotion and stress detection tasks, integrating physiological and behavioral data to improve generalization. Methodological surveys~\cite{fukazawa2020smartphone,yang2022survey,vos2023generalizable} have since summarized the common pipeline for mobile and wearable emotion sensing, typically encompassing (1) data collection and preprocessing, (2) feature extraction, and (3) model training and evaluation using ML or DL methods.

Finally, researchers have begun to address the generalizability challenge; whether models trained on one cohort or dataset can perform well on another. \citet{xu2023globem} examined the cross-dataset robustness of depression detection models across institutions and years, while others explored transfer learning and data-merging strategies for stress, mood, and personality inference~\cite{meegahapola2023generalization,khwaja2019modeling,adler2022machine}. These studies underscore a persistent methodological gap: existing models often lack cross-user generalization, motivating new research directions that leverage large, foundation models, such as LLMs, to unify sensing, reasoning, and personalization in mental health.

\subsection{Machine/Deep Learning for Mental Health Forecasting Using Mobile Sensing Data}

Recent studies have explored forecasting mental health outcomes such as mood, stress, and depression using passive mobile sensing data. \citet{umematsu2020forecasting} used wrist-worn physiological sensors to forecast next-day stress, mood, and health levels among office workers and students, showing that physiological signals can moderately predict well-being. \citet{langener2024predicting} combined experience sampling, digital phenotyping, and social network data to model mood in daily life, finding that social-context features contribute to prediction but performance varies greatly across individuals.

From a modeling standpoint, \citet{spathis2019sequence} introduced a sequence multi-task recurrent neural network to forecast mental well-being from sparse self-reports, showing that temporal and multi-task learning improve predictive accuracy. \citet{kathan2022personalised} extended forecasting to mobile sensor data and ecological momentary assessments (EMAs), demonstrating that personalized models outperform global ones in predicting depression severity. \citet{schat2024forecasting} compared person-specific and multilevel exponentially weighted moving average models to forecast depression onset with minimal baseline data, highlighting the tradeoff between personalization and generalization. \citet{paz2025emotion} further introduced a transformer-based framework for emotion forecasting, achieving high accuracy through attention-based temporal reasoning. 

Overall, prior work has progressed from correlation analyses to advanced temporal and personalized modeling. Yet, most studies focus on narrow sensing modalities or specific tasks. No prior work has systematically compared traditional ML, deep learning, and LLM approaches for mental health forecasting using longitudinal passive sensing data, which is the gap our study addresses.
\section{Dataset Overview and Analysis}

In this section, we first overview the College Experience Sensing (CES) dataset used in this study, and then present a preliminary analysis to demonstrate why it is well-suited for investigating mental health forecasting.

\subsection{Dataset Overview}\label{subsec:dataset}
Our study leverages the College Experience Sensing (CES) dataset~\cite{nepal2024capturing}, the longest publicly available longitudinal mobile sensing dataset centered on college students’ mental health. Released by Dartmouth College in October 2024, CES is particularly valuable for its five-year continuous data collection (2017–2022), encompassing the pre-pandemic, pandemic, and post-pandemic recovery phases. This rich temporal coverage offers a unique opportunity to examine behavioral and mental health dynamics of college students across distinct societal and environmental contexts.
The dataset includes 215 college students from Dartmouth College who installed the application on their personal smartphones. Data was collected through two primary channels: (1) \textit{\textbf{passive sensing data}}, which continuously captured daily behavioral traces from phone sensors on an hourly basis, including mobility, physical activity, sleep patterns, and phone usage; and (2) \textit{\textbf{self-report data}}, gathered via Ecological Momentary Assessment (EMA) surveys sent weekly to participants to collect mental health outcomes. This combination of objective behavioral data and subjective self-reports in a longitudinal format provides a robust foundation for building accurate mental health forecasting models. 

The \textit{passive sensing data} is high-dimensional, comprising 172 original features that capture a broad spectrum of daily behavioral patterns. Directly utilizing all features would not only incur significant computational overhead but also risk overfitting and reduced generalizability due to redundancy and noise. To address this, prior work I-HOPE~\cite{roychowdhury2025predicting} employs a systematic feature selection process to obtain a refined subset of \textit{\textbf{35 behavioral features per day}}, effectively representing key aspects of daily life such as sleep rhythms, social interactions, phone usage habits, and activity patterns.
The mental health state labels in our study are derived from the Patient Health Questionnaire-4 (PHQ-4) scores~\cite{kroenke2009ultra} collected through participants' \textit{self-report data}. The PHQ-4 is a well-validated screening instrument designed to assess anxiety and depression based on participants' experiences over the previous two weeks. Its total score ranges from 0 to 12, with higher values indicating greater symptom severity. Following the categorization adopted in the I-HOPE study, we discretize the continuous PHQ-4 scores into \textit{\textbf{four clinically meaningful severity levels}}: Normal (0–3), Mild (4–6), Moderate (7–9), and Severe (10–12). This transformation not only simplifies the model's prediction task but also enhances the clinical interpretability and actionability of the results for potential mental health intervention.


\begin{table}[t]
    \centering
    \footnotesize
    \setlength{\tabcolsep}{3.5pt}
    \caption{\textbf{Statistics of Dataset.} `\#Min\_/Max\_/Avg\_Sam' indicates the minimum, maximum, and average number of samples per participant; and the last four columns indicate the number of samples per level.}
    \vspace{-0.1in}
    \begin{tabular}{c|c|c|c|c|c|c|c|c|c}
    \toprule
    \# Students &\#Features  &\#Total\_Sam &\#Min\_Sam &\#Max\_Sam &\#Avg\_Sam  &\#Normal\_Sam &\#Mild\_Sam &\#Moderate\_Sam &\#Severe\_Sam \\\midrule
    215     &35 &24,778 &61 &432 &206 &15,477 &6,524 &1,795 &982  \\
    \bottomrule
    \end{tabular}
    \label{tab:statistics}
\end{table}

Accordingly, in our study, each data sample comprises \textit{\textbf{two weeks of passive sensing data}} paired with the corresponding \textit{\textbf{mental health severity level}} derived from the PHQ-4 score. A summary of the dataset's statistics is presented in Table~\ref{tab:statistics}.

\subsection{Dataset Analysis}

We conduct a preliminary analysis to better understand the dataset's structure and characteristics at both the population and individual levels, which helps reveal overall behavioral and mental health state trends for downstream forecasting modeling.

\subsubsection{Analysis of Population Difference}\label{subsubsec:analysis_population}
We first examine whether the 35 core behavioral features from different mental health levels (classes) exhibit statistically distinguishable patterns. To assess this, we analyze the similarity of data samples within and across classes. Specifically, we first flatten each 35-dimensional daily feature record into a single vector and then concatenate all daily vectors within a given PHQ-4 assessment period (i.e., two weeks) to represent the user’s overall behavioral state. 
We then compute pairwise cosine similarities among all vectors, both within and across the four classes. Fig.~\ref{fig:data_analysis}(a) shows the results, revealing two key observations. First, \textit{\textbf{the three abnormal classes (Mild, Moderate, and Severe) all exhibit relatively low similarity with the Normal class}}, indicating that deviations in behavior between normal and abnormal mental states are well captured by the features. Second, \textit{\textbf{within the abnormal classes, the boundaries are less distinct, particularly between Mild and Severe}}, indicating overlapping behavioral patterns among these intermediate states. Although the Severe class exhibits relatively good separability with the highest intra-class similarity (0.0405), it also shows a high inter-class similarity with the Mild class (0.0405), suggesting behavioral convergence between the two groups.
These observations highlight the inherent challenge of accurately classifying different abnormal cases, making them critical benchmarks for evaluating model performance in subsequent analyses.

\subsubsection{Analysis of Individual Difference}\label{subsub:individual-analysis}
\begin{figure*}[t]
\centering
\graphicspath{{figure/}} 
\begin{minipage}{0.35\linewidth}
  \centerline{\includegraphics[width=7cm]{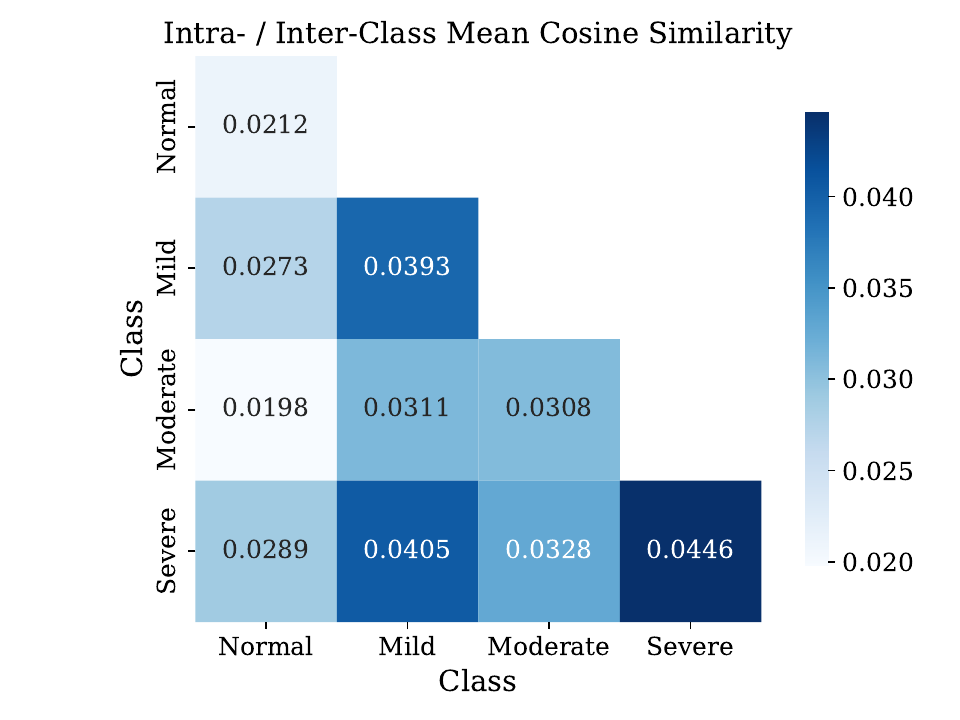}}
\end{minipage}
\hspace{0.3in}
\begin{minipage}{0.5\linewidth}
  \centerline{\includegraphics[width=7cm]{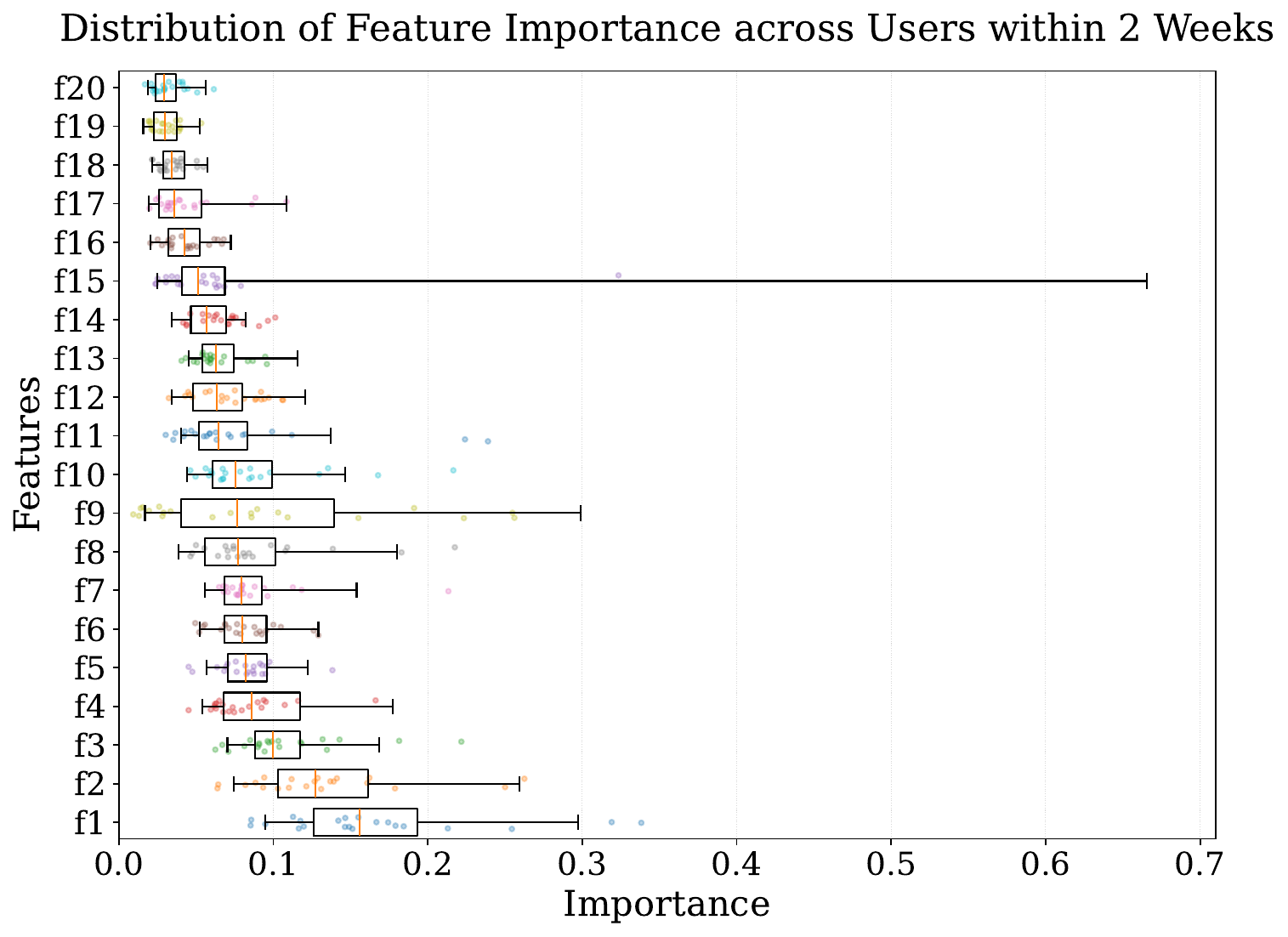}}
\end{minipage}
\caption{\textbf{(a) Intra and inter-class similarity of behavior patterns; (b) The importance of top-20 features across individual users} (`f1: sleep\_duration', `f2: act\_running', `f3: 'unlock\_num/unlock\_duration', `f4: act\_walking', `f5: loc\_home\_dur', `f6: act\_on\_bike', `f7: act\_still', `f8: loc\_home\_unlock\_num/loc\_home\_unlock\_duration', `f9: loc\_home\_audio\_voice', `f10: loc\_workout\_dur', `f11: sleep\_start', `f12: loc\_other\_dorm\_dur', `f13: sleep\_end', `f14: loc\_study\_unlock\_num/loc\_study\_unlock\_duration', `f15: act\_on\_foot', `f16: loc\_leisure\_dur', `f17: loc\_study\_dur', `f18: loc\_social\_unlock\_num/loc\_social\_unlock\_duration', `f19: loc\_other\_dorm\_unlock\_num/loc\_other\_dorm\_unlock\_duration', `f20: loc\_self\_dorm\_dur').
}
\label{fig:data_analysis}
\end{figure*}

We aim to further examine whether the 35 core behavioral features exert a uniform influence across individuals. To investigate this, we conduct a feature importance analysis using XGBoost~\cite{chen2016xgboost} for all participants. We choose XGBoost as it can quantify each feature's contribution to model performance by measuring how often and how effectively it splits data to reduce prediction error across all decision trees. The results, visualized in the box plot in Fig.~\ref{fig:data_analysis} (b), reveal two key insights. Note that due to space limitations, we only show the importance of the top-20 features.  
First, \textit{\textbf{many features exhibit varying levels of importance across individuals}}. For example, the importance of `f15: act\_on\_foot' (activity on foot) ranges from 0.018 to 0.9819, while that of `f9: loc\_home\_audio\_voice' (the average proportion of time with detectable activity or sound while at home) varies between 0.03 and 0.3764.
Second, \textit{\textbf{some features show relatively high average importance among participants but still display considerable inter-individual variability}}. For instance, the median importance of `f1: sleep\_duration' and `f2: act\_running' (time spent running) are 0.1715 and 0.1581, respectively - both higher than that of `f20: loc\_self\_dorm\_dur' (time spent at own dorm), which has a median importance of 0.0525. However, the importance of `f1: sleep\_duration' and `f2: act\_running' spans a wide range, from 0.0781 to 0.3948 and 0.0088 to 0.3541, respectively.
Third, \textit{\textbf{some features demonstrate relatively consistent impacts across individuals}}. For example, the importance values of `f20: loc\_self\_dorm\_dur' are tightly clustered, with a maximum of 0.1086, a median of 0.0525, and a minimum of 0.0107. 
In summary, these observations provide a strong motivation for exploring models capable of capturing personalized information. 

\section{Methodology}
Our methodology is designed to systematically address the six research questions (RQs) presented in Section~\ref{sec:intro}, focusing on mental health forecasting using smartphone sensing data. In this section, we first formally define the forecasting task, highlighting its distinction from mental health detection. We then describe the feature representation strategies applied to prepare the sensing data, followed by an overview of the three modeling approaches explored in this study: Traditional Machine Learning (ML), Deep Learning (DL), and LLMs. Finally, we describe the personalization modeling techniques and the class imbalance mitigation strategies incorporated to enhance model robustness and fairness.

\subsection{Problem Formulation}
To establish a clear basis for our comparative study, we first distinguish between two related but distinct tasks: mental health detection vs. mental health forecasting. This distinction is critical as \textit{\textbf{forecasting}}, \textit{\textbf{our primary focus}}, directly addresses the proactive goal of anticipating mental health changes before they manifest, a key challenge highlighted in the introduction.

\subsubsection{Mental Health Detection}

The task of mental health detection is formally defined as a multi-class time-series classification problem. Given a sequence of smartphone sensing data for a user $u$ over a period of $T$ days, $\mathbf{X}_{1:T}^{(u)} = (\mathbf{x}_1, \mathbf{x}_2, \dots, \mathbf{x}_T)$, where $\mathbf{x}_t \in \mathbb{R}^D$ is the feature vector for day $t$ and $D$ is the feature dimension, the objective is to learn a function $f_{\mathrm{det}}$ that predicts the corresponding mental health state $y_T^{(u)} \in \{C_1, C_2, \dots, C_K\}$:
\begin{equation}
\hat{y}_T^{(u)} = f_{\mathrm{det}}\left( \mathbf{X}_{1:T}^{(u)} \right),
\end{equation}
where $K$ is the number of distinct mental health states. Detection is a reactive task, aiming to infer the user’s present mental condition once behavioral symptoms have already emerged.

\subsubsection{Mental Health Forecasting} The task of mental health forecasting extends beyond detection by aiming to predict future mental health states before symptoms become apparent. Formally, given the same sequence of past observations $\mathbf{X}_{1:T}^{(u)}$, the model learns a mapping function $f_{\mathrm{fore}}$ that forecasts the mental health state label at a future time step $T + \Delta$ (e.g., next day or next week):

\begin{equation}\label{eq:forecast} 
\hat{y}^{(u)}_{T+\Delta} = f_{\mathrm{fore}}\!\left( \mathbf{X}^{(u)}_{1:T} \right),
\end{equation}
where $\Delta > 0$ denotes the forecast horizon. Forecasting is therefore a proactive task that can anticipate and prevent mental health deterioration before critical episodes occur.

\subsection{Feature Representation Strategies}\label{subsec:feature_representation}

Before training any forecasting models, the raw, high-dimensional sensing data must be transformed into a structured representation that is both semantically meaningful and computationally tractable. To this end, we design feature representation strategies that explore different levels of granularity in terms of feature dimensionality and temporal duration. The effects of these design choices are systematically evaluated in Section~\ref{subsec:feature_rep_results}. 
  
\begin{figure}
    \centering
    \includegraphics[width=1\linewidth]{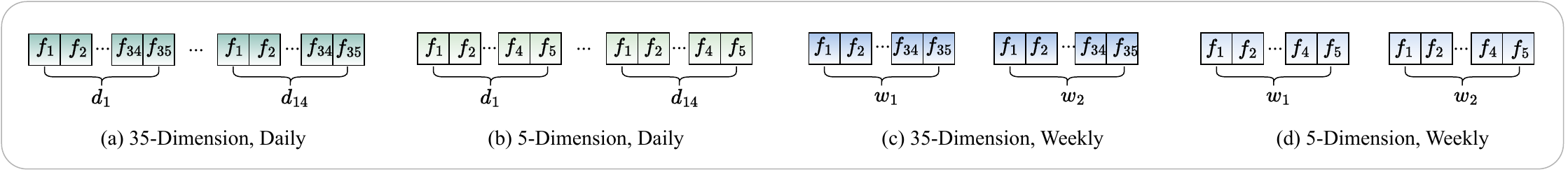}
    \caption{Four distinct feature representation configurations.}
    \label{fig:feature_rep}
\end{figure}

To examine the effect of \textit{\textbf{feature dimension}} ($D$) on forecasting performance, we adopt two feature organization strategies following I-HOPE~\cite{roychowdhury2025predicting}:
(1) \textbf{35-Dimension}, which directly utilizes the refined set of 35 core behavioral features (details available in~\cite{roychowdhury2025predicting}); and
(2) \textbf{5-Dimension}, which aggregates these 35 features into five broader behavioral categories that capture major aspects of daily life - leisure, me time, phone time, sleep, and social time.
Second, to assess the impact of \textit{\textbf{temporal duration}} ($T$), we consider two temporal granularity settings:
(1) \textbf{Daily}, where each day’s data is treated as an individual time step to capture short-term behavioral fluctuations and transitions; and
(2) \textbf{Weekly}, where daily observations are averaged within each week to reduce short-term noise and highlight longer-term behavioral trends. Thus, four distinct feature representation configurations as shown in Fig.~\ref{fig:feature_rep} are considered, reflecting different combinations of feature dimension and temporal duration: (35-Dimension, Daily), (35-Dimension, Weekly), (5-Dimension, Daily), and (5-Dimension, Weekly).

\subsection{Modeling Approaches}
In our study, we systematically evaluate three types of distinct modeling approaches for the mental health forecasting task, including traditional machine learning (ML), deep learning (DL), and LLMs.  

\subsubsection{Traditional ML Approaches}
\label{subsec:traditional-ml}
For each data sample, we apply two strategies to transform its two-week time-series sensing data into fixed-length input representations:
(1) Statistical Aggregation: the feature values of each dimension (35-Dimension or 5-Dimension) over the two-week period are averaged to summarize overall user behavior. This approach reduces noise and captures general behavioral trends, which is appropriate for models that cannot inherently model temporal dependencies; and
(2) Sequential Flattening: the feature vectors (Daily or Weekly) are concatenated in chronological order to form a single high-dimensional vector. This method retains temporal context and preserves the sequence of behaviors, albeit at the cost of increased feature dimensionality~\footnote{We empirically find the Sequential Flattening strategy performs significantly worse than Statistical Aggregation for traditional ML. This performance gap may be attributed to the fact that traditional ML models cannot inherently capture temporal dependencies.}.
Meanwhile, 
we evaluate five representative traditional ML models, each taking the processed input and predicting a categorical mental health label as output. 
\begin{itemize}[leftmargin=*]
    \item \textit{Logistic Regression}~\cite{hosmer2013applied,chikersal2021detecting}: A linear model that estimates class probabilities through a logistic function, mapping behavioral features directly to mental health classes.
    \item \textit{Support Vector Machine (SVM)}~\cite{brereton2010support,zhu2023stress}: A margin-based classifier that finds an optimal hyperplane separating different mental health classes in the feature space.
    \item \textit{Decision Trees}~\cite{song2015decision,opoku2021predicting}: A tree-structured model that recursively splits features based on information gain, producing interpretable decision rules for mental health classification.
    \item \textit{Random Forests}~\cite{breiman2001random,lewis2023mixed}: An ensemble of multiple decision trees trained on random subsets of data and features, reducing overfitting and improving robustness.
    \item \textit{Gradient Boosting (XGBoost~\cite{chen2016xgboost}, LightGBM~\cite{ke2017lightgbm})}~\cite{shen2025passive}: A boosting-based ensemble that builds trees sequentially, where each new tree corrects the residual errors of the previous ones to achieve high predictive accuracy.
\end{itemize}

\subsubsection{Deep Learning Approaches}
\label{subsec:deep-learning}
Unlike traditional ML models, DL models can directly process sequential data and capture temporal behavioral dynamics for sequence modeling. Accordingly, the input is represented as an ordered sequence of feature vectors (Daily or Weekly): $\mathbf{X}^{(u)} = (\mathbf{x}_1, \mathbf{x}_2, \dots, \mathbf{x}_T)$. We then evaluate four representative DL models: 
\begin{itemize}[leftmargin=*]
    \item \textit{Multi-Layer Perceptron (MLP, I-HOPE~\cite{roychowdhury2025predicting}):} Processes aggregated feature vectors without modeling sequential order, which serves as a baseline for evaluating the advantages of sequential models.
    \item \textit{Temporal Convolutional Networks (TCN)}~\cite{bai2018empirical}: Uses causal convolutions to capture local temporal dependencies, assuming short-term behavioral changes are informative of mental state variation.
    \item \textit{{Long Short-Term Memory (LSTM)} + Attention}~\cite{umematsu2019daytime, yu2019review}: Models long-range temporal dependencies through gated recurrent mechanisms, suitable for capturing evolving behavioral patterns, where the attention mechanism is applied to distinguish the importance of features at different time steps. 
    \item \textit{Transformer}~\cite{vaswani2017attention,paz2025emotion}: Learns attention-weighted temporal relationships, enabling the model to focus on relevant time periods dynamically. 
\end{itemize}

\subsubsection{Large Language Model Approaches}
\label{subsec:llm}
LLMs provide a new paradigm for learning patterns from behavioral time-series data. To make sensor data interpretable to LLMs, we transform the numerical features $\mathbf{X}^{(u)}$ into structured text using the following three steps, enabling LLMs to reason about behaviors linguistically. 
\begin{itemize}[leftmargin=*]
    \item \textit{Feature Renaming:} Each sensing variable is renamed into a human-readable term that LLMs can easily interpret, e.g., “screen\_on\_time (minutes)". Units are explicitly added to preserve quantitative meaning.
    \item \textit{Temporal Ordering:} Behavioral records are arranged chronologically on a Daily or Weekly basis (e.g., Day 1, Day 2, …), ensuring that temporal dependencies are reflected in the text sequence.
    \item \textit{Structured Formatting:} The data are formatted as a markdown table, where each row represents a day (week) and each column represents a feature. The cell values contain the actual feature values. This tabular structure helps LLMs parse temporal and contextual information in a consistent, interpretable way.
\end{itemize}
In summary, this representation bridges numeric sensor data and natural language understanding, enabling LLMs to reason over user behavior using their pretrained contextual and temporal knowledge. 

To better align the pre-trained knowledge of LLMs with the behavioral-driven mental health forecasting objective, we implement \textbf{three adaptation strategies} as follows, namely zero-shot, in-context learning, and parameter efficient fine-tuning. 

\begin{itemize}[leftmargin=*]
    \item \textit{Zero-shot}~\cite{xu2024mental}: The LLM purely receives the serialized behavioral text and task description without any examples. It relies solely on its pretrained knowledge to infer the relationship between behaviors and mental states.
    \item \textit{In-context Learning} (ICL)~\cite{kim2024healthllm}: The LLM is guided by additional reference information embedded in the prompt to improve its task understanding. The contextual knowledge can take several complementary forms:
    \begin{itemize}
        \item \textit{Few-shot Learning}: The LLM is provided with a few labeled examples identical in format to the target forecasting task. These examples can be selected from: (1) Recency-based examples, drawn from the most recent behavioral data to reflect current user mental health states; or (2) Similarity-based examples, retrieved from historical records that resemble the target behavioral pattern.
        \item \textit{Statistical Knowledge}: For each mental health class, we compute statistical summaries, including the mean and variance of all behavioral features at either the \textit{individual} or \textit{population} levels. These statistical descriptors are then provided to the LLM as quantitative priors, enabling it to interpret the behavioral patterns within the target forecasting task. 
        \item \textit{Pattern Knowledge}: Behavioral data and their corresponding labels for each mental health class are provided to a more capable LLM (e.g., GPT-4.1), which summarizes and abstracts the distinguishing behavioral patterns across classes. The derived patterns are then used as structured knowledge to support reasoning and prediction in the target forecasting task~\footnote{We empirically observe in Section~\ref{subsec:llm_variants} that individual-level knowledge is more effective than population-level knowledge (e.g., Statistical\_individual outperforms Statistical\_population) in Table~\ref{tab:llm-ablation}. Therefore, we summarize and incorporate personalized behavioral patterns as contextual information in the Pattern Knowledge strategy.}.
    \end{itemize}
    \item \textit{Parameter-Efficient Fine-Tuning (PEFT)}~\cite{han2024parameter}: To adapt LLMs to the mental health forecasting task, we first construct input–output pairs specifically designed for the target objective. The backbone LLM is then fine-tuned by updating only a small subset of additional parameters, while keeping the backbone LLM parameters frozen. This approach maintains the generalization capability of the backbone LLM while enabling efficient and domain-specific adaptation. We consider two representative PEFT techniques:
    \begin{itemize}
        \item \textit{Low-Rank Adaptation (LoRA)}~\cite{dettmers2023qlora}: Inserts low-rank trainable matrices into specific layers of the backbone LLM, such as attention layers, to capture task-specific information with minimal parameter updates, achieving efficient task adaptation.
        \item \textit{Prompt Tuning}~\cite{liliang2021prefix}: Learns a small set of trainable \textit{soft prompts} that serve as continuous embeddings of the behavioral features prepended to the input of LLMs. Specifically, we first pre-train a Transformer encoder to model user behavioral sequences. During pre-training, a random subset of feature values is masked, and the encoder is trained with a regression head to reconstruct the masked values. This masked reconstruction objective enables the encoder to learn robust temporal dynamics and cross-feature dependencies. Next, we introduce a lightweight \textit{projector} that maps the encoded behavioral embeddings into the semantic space of the LLM. Thus, the projected embeddings serve as continuous \textit{soft prompts}, which are concatenated with textual task instructions and fed into the frozen LLM. Through this alignment mechanism, the LLM can directly interpret and reason over pre-trained time-series representations without explicit textualization of behavioral data, achieving efficient and semantically grounded task adaptation.
    \end{itemize}
\end{itemize}

\subsection{Personalization Modeling}
As discussed in Section~\ref{subsub:individual-analysis}, individuals often exhibit distinct behavioral patterns, and the importance of behavioral features can vary substantially across participants. This observation strongly motivates the exploration of models capable of capturing personalized behavioral variations. Therefore, for each modeling approach, we consider two operational modes: 
\begin{itemize}[leftmargin=*]
    \item \textbf{User-agnostic Mode:} Learns population-level patterns without explicit user identification, assuming that behavioral–mental health relationships generalize across individuals.
    \item \textbf{User-aware Mode:} Incorporates user identity to enable the model to capture individual-specific and personalized behavior patterns. 
    \begin{itemize}
        \item For \textbf{Traditional ML models}, user identity is introduced by adding a one-hot encoded \textit{user ID} as an additional categorical feature alongside behavioral variables.
        \item For \textbf{DL models}, personalization is achieved through a user embedding mechanism, where each \textit{user ID} is mapped to a learnable low-dimensional vector that identifies their unique behavioral tendencies. This embedding is concatenated with the feature vector at each time step, enabling the model to personalize its predictions while simultaneously learning shared temporal dynamics across all users.
        \item For \textbf{LLMs}, the \textit{user ID} is textualized and inserted into the prompt as a personalized identifier of behavioral data. This approach is only applied under the PEFT setting, as effective association between user identity and behavioral patterns can only be learned during the training or fine-tuning process.
    \end{itemize}
\end{itemize}

\subsection{Class Imbalance}\label{subsec:med_class_imbalance}
As shown in Table~\ref{tab:statistics}, the data distribution across mental health classes is notably imbalanced. For instance, the \textit{Normal} class contains 15,477 samples (62\% of the total), whereas the \textit{Moderate} class includes only 1795 samples (7\% of the total). Such an imbalance can adversely affect model training, as models may become biased toward the majority class or tend to predict all samples as belonging to it to achieve deceptively high accuracy. This phenomenon has been empirically examined in Sections~\ref{subsec:overall_res}-\ref{subsec:class_imbalance} (referring to the imbalanced F1 score across different classes).
To address the issue of class imbalance, we investigate the effectiveness of different loss functions designed to reweight or refocus learning on minority classes:
\begin{itemize}[leftmargin=*]
    \item \textit{Weighted Cross-Entropy Loss}~\cite{aurelio2019learning}: Assigns higher weights to minority classes based on their inverse frequency, encouraging the model to pay greater attention to underrepresented samples.
    \item \textit{Focal Loss}~\cite{lin2017focal}: Further refines this by down-weighting easy examples and concentrating training on harder, misclassified cases, thereby improving robustness under severe imbalance conditions.
\end{itemize}

\section{Experimental Setup, Results, and Analysis}\label{sec:experiment}

In this section, we first describe the experimental setup, then present numerical results addressing each research question, and finally provide detailed analyses to interpret the findings.

\subsection{Experimental Setup}
\label{subsec:setup}

\subsubsection{Dataset} We use the College Experience Sensing (CES) dataset~\cite{nepal2024capturing}, which is described in detail in Section~\ref{subsec:dataset}, and its summary statistics are presented in Table~\ref{tab:statistics}. To prevent data leakage across splits, the dataset is partitioned at the user level into training, validation, and test sets following a 7:1:2 ratio. Specifically, for each user, the earliest 70\% of samples are used for training, the subsequent 10\% for validation, and the final 20\% for testing.

\subsubsection{Evaluation Metrics} Following prior work in mental health detection and forecasting~\cite{roychowdhury2025predicting,chikersal2021detecting}, we report Precision, Recall, and F1-score for each class to evaluate classification performance, along with Accuracy and Macro-F1 as overall metrics reflecting performance across all classes. 

\subsubsection{Implementation Details}
\label{subsubsec:env-impl}
Traditional ML models are implemented using scikit-learn~\cite{scikit-learn}, while DL models are implemented with PyTorch~\cite{paszke2019pytorch}. For LLMs, we employ both open-source Qwen models (4B, 8B, 14B)~\cite{qwen2023technical} and the commercial GPT-4.1 model~\cite{openai_gpt4.1_2025} for zero-shot and in-context learning. Furthermore, we apply PEFT techniques, including LoRA and prompt tuning, implemented via the HuggingFace Transformers~\cite{wolf-etal-2020-transformers} framework on Qwen models. All experiments are conducted on NVIDIA A100 GPU with 40GB memory. We empirically determine the optimal parameter settings for all ML, DL models, and LLMs with PEFT based on their performance on the validation set. For all methods, during training, each sample uses the two-week time-series features as input to predict the corresponding mental health level. In contrast, during testing, only the first-week time-series features are provided as input to predict the mental health level in the second week (i.e. $T=7 \text{ days}$ and $\Delta=7 \text{ days}$ in Equation~\ref{eq:forecast}), consistent with our focus on the forecasting task. The implementation details and source code are provided in our Git repository, with the link included in Section~\ref{subsec:llm}.  

\subsection{Comparison Across Modeling Approaches (RQ1)}\label{subsec:overall_res}

We first compare the overall performance of the three modeling approaches---Traditional ML, DL, and LLMs---across all evaluation metrics for the mental health forecasting task. Each model is optimized under its best configuration, considering feature representation, personalization modeling, and loss function. Empirically, we find that all methods perform better when incorporating personalization, and most achieve stronger results when combined with weighted or focal loss and coarse-grained feature representations. 
Regarding LLMs, we also select their best adaptation strategies, and we empirically find that few-shot learning consistently yields the best performance across all LLMs. The results are summarized in Table~\ref{tab:overall}, where the best performance within each category is highlighted in \textbf{bold}, and the overall best performance across all categories is shaded in \textcolor{blue!80}{blue}.
Several key observations can be drawn from these results.

\begin{table}[t]
\centering
\footnotesize
\setlength{\tabcolsep}{1.7pt}
\caption{\textbf{Performance Comparison Across Different Approaches and Severity Levels (RQ1, \ref{subsec:overall_res}).} The best performance within each category is highlighted in \textbf{bold}, and the overall best performance across all categories is shaded in \textcolor{blue!80}{blue}.}
\label{tab:overall}
\vspace{-0.1in}
\begin{tabular}{ll|ccc|ccc|ccc|ccc|cc}
\toprule
\multirow{2}{*}{} & \multirow{2}{*}{Model} & \multicolumn{3}{c|}{Normal} & \multicolumn{3}{c|}{Mild} & \multicolumn{3}{c|}{Moderate} & \multicolumn{3}{c|}{Severe} & \multirow{2}{*}{Acc} & \multirow{2}{*}{Macro-F1} \\
& & Pre & Rec & F1 & Pre & Rec & F1 & Pre & Rec & F1 & Pre & Rec & F1 & & \\
\midrule
\multirow{6}{*}{\begin{tabular}[c]{@{}l@{}}ML\end{tabular}}
& Random Forest & 0.6212 & 0.9755 & 0.7590 & \cellcolor{blue!20}\textbf{0.5774} & 0.1934 & 0.2898 & 0.2609 & 0.0274 & 0.0496 & \cellcolor{blue!20}\textbf{0.9451} & 0.3613 & \textbf{0.5228} & 0.6190 & 0.4053 \\
& Decision Tree & 0.5737 & \cellcolor{blue!20}\textbf{0.9909} & 0.7267 & 0.5180 & 0.0439 & 0.0810 & 0.1351 & 0.0114 & 0.0211 & 0.6452 & 0.1681 & 0.2667 & 0.5699 & 0.2738 \\
& XGBoost & 0.6571 & 0.9513 & 0.7773 & 0.5737 & 0.2874 & 0.3829 & \textbf{0.3529} & 0.0685 & 0.1147 & 0.7077 & 0.3866 & 0.5000 & 0.6401 & 0.4437 \\
& LightGBM & 0.6874 & 0.9247 & \textbf{0.7886} & 0.5469 & 0.3344 & 0.4150 & 0.3164 & 0.1279 & 0.1821 & 0.5533 & 0.3487 & 0.4278 & 0.6436 & 0.4534 \\
& SVM & 0.7075 & 0.8894 & 0.7881 & 0.5543 & \textbf{0.4265} & \textbf{0.4821} & 0.3354 & 0.1256 & 0.1827 & 0.6474 & 0.4244 & 0.5127 & \textbf{0.6565} & 0.4914 \\
& Logistic Regression & \textbf{0.7905} & 0.7546 & 0.7721 & 0.5244 & 0.4198 & 0.4663 & 0.2826 & \textbf{0.4110} & \textbf{0.3349} & 0.3165 & \cellcolor{blue!20}\textbf{0.6597} & 0.4278 & 0.6150 & \textbf{0.5003} \\
\hline
\multirow{4}{*}{DL}
& MLP (I-HOPE~\cite{roychowdhury2025predicting}) & 0.7223 & \textbf{0.8708} & \cellcolor{blue!20}\textbf{0.7896} & \textbf{0.5477} & 0.4655 & 0.5033 & \cellcolor{blue!20}\textbf{0.3750} & 0.1781 & 0.2415 & \textbf{0.9213} & 0.4916 & \textbf{0.6411} & \cellcolor{blue!20}\textbf{0.6662} & 0.5439 \\
& TCN & 0.7333 & 0.8396 & 0.7829 & 0.5344 & 0.4308 & 0.4770 & 0.3600 & 0.3288 & 0.3437 & 0.6722 & \textbf{0.5084} & 0.5789 & 0.6515 & 0.5456 \\
& LSTM+Attention & 0.7617 & 0.7735 & 0.7675 & 0.5116 & \cellcolor{blue!20}\textbf{0.5674} & \cellcolor{blue!20}\textbf{0.5380} & 0.2878 & 0.2215 & 0.2503 & 0.7672 & 0.3739 & 0.5028 & 0.6430 & 0.5147 \\
& Transformer & \cellcolor{blue!20}\textbf{0.8206} & 0.7335 & 0.7746 & 0.5130 & 0.5284 & 0.5206 & 0.2989 & \cellcolor{blue!20}\textbf{0.5479} & \cellcolor{blue!20}\textbf{0.3868} & \textbf{0.9213} & 0.4916 & \cellcolor{blue!20}\textbf{0.6411} & 0.6416 & \cellcolor{blue!20}\textbf{0.5808} \\
\hline
\multirow{4}{*}{LLM}
& Qwen3-4B & \textbf{0.6935} & 0.8151 & 0.7494 & 0.4856 & 0.3490 & 0.4061 & 0.2014 & 0.1918 & 0.1965 & 0.4338 & \textbf{0.3992} & \textbf{0.4158} & 0.5954 & \textbf{0.4419} \\
& Qwen3-8B & 0.6920 & 0.8151 & 0.7486 & 0.4833 & \textbf{0.3527} & 0.4078 & 0.2078 & 0.1941 & 0.2007 & \textbf{0.4455} & 0.3782 & 0.4091 & 0.5958 & 0.4415 \\
& Qwen3-14B & 0.6926 & 0.8197 & 0.7508 & 0.4897 & 0.3496 & \textbf{0.4080} & 0.2024 & 0.1918 & 0.1970 & 0.4417 & 0.3824 & 0.4099 & \textbf{0.5974} & 0.4414 \\
& GPT-4.1 & 0.6902 & \textbf{0.8246} & \textbf{0.7514} & \textbf{0.4907} & 0.3368 & 0.3994 & \textbf{0.2094} & \textbf{0.2032} & \textbf{0.2063} & 0.4354 & 0.3824 & 0.4072 & 0.5970 & 0.4411 \\
\bottomrule
\end{tabular}
\end{table}

(1) \textit{\textbf{DL models achieve the highest overall performance.}} 
Among all modeling approaches, DL models consistently outperform both traditional ML models and LLMs, achieving the best macro-F1 score (0.5808 by Transformer). In particular, Transformer and MLP demonstrate strong and balanced performance across all severity levels, reflecting their ability to capture temporal dependencies and nonlinear behavioral patterns.

(2) \textit{\textbf{Traditional ML models show competitive results for common classes but struggle with rare ones}}. ML models such as SVM and Logistic Regression perform relatively well for the Normal class, with F1-scores above 0.77, but their performance drops sharply for Mild and Moderate cases. This disparity highlights their limitation in handling class imbalance and complex temporal dynamics in behavioral data.

(3) \textit{\textbf{LLMs exhibit moderate performance and limited generalization across classes}}.
Despite their strong reasoning capabilities, LLMs (Qwen3 series and GPT-4.1) show overall Macro-F1 scores around 0.44, which are lower than those of their DL counterparts and even most traditional ML models (e.g., SVM). Their extensive knowledge and powerful reasoning capability do not yet translate to superior predictive accuracy. The possible reason is that they are pre-trained on large-scale general text corpora and thus lack domain-specific knowledge essential for specialized applications. Even with PEFT for task adaptation, LLMs remain constrained by their reliance on textual sequence representations, which are insufficient to comprehensively model user behavior or classification patterns derived from historical and structured numerical sensing data. 
Besides, \textit{\textbf{there is very little performance difference among all LLMs}}. Qwen3 models of varying sizes (4B, 8B, 14B) and GPT-4.1 all yield nearly identical Macro-F1 scores, around 0.441. This indicates a performance plateau, where increasing model size does not lead to better results for this specific forecasting task.

(4) \textit{\textbf{Across all models, the Normal class consistently achieves substantially higher F1 score compared to the rest classes}}. This pattern highlights the inherent difficulty of forecasting rare, high-severity mental health states, which aligns with the class imbalance discussed earlier in Section~\ref{subsec:med_class_imbalance}. 
Notably, the F1 score of the Severe class ranks second, despite having the fewest samples. This is likely because it exhibits the highest intra-class similarity (0.0446), indicating relatively good separability, as analyzed in Section~\ref{subsubsec:analysis_population}. However, it also shows a high inter-class similarity with the Mild class (0.0405), suggesting overlapping behavioral patterns between these two groups. Moreover, almost all models exhibit the lowest F1 score on the Mild and Moderate classes. This can be attributed to the relatively high behavioral similarity between individuals in these two categories, which leads to less distinct class boundaries and greater classification ambiguity, as discussed in Section~\ref{subsubsec:analysis_population}.


\subsection{Comparison Across Different LLM Variants (RQ2)}\label{subsec:llm_variants}

We further examine how different LLM adaptation strategies affect the forecasting performance, including Zero-shot learning, four types of In-context Learning (i.e. Similarity-, Recency-, Statistical-, and Pattern-based) and two PEFT strategies (i.e. LoRA and Prompt Tuning) as introduced in Section~\ref{subsec:llm}.
It is important to note that: (1) We report results only for the (35-Dimension, Daily) configuration, as similar trends are noted across other feature granularities; (2) PEFT strategies could not be applied to the source-closed GPT-4.1 model; and (3) Prompt Tuning empirically underperforms LoRA on Qwen3-8B, and due to time constraints, we did not extend this comparison to other Qwen variants.
Results are summarized in Table~\ref{tab:llm-ablation}, where the best performance within each category on each metric is highlighted in bold, and the overall best performance on each metric across all categories is shaded in blue.
Some observations can be concluded. 

\begin{table}[t]
\centering
\footnotesize
\setlength{\tabcolsep}{1.8pt}
\caption{\textbf{Performance Comparison across Different Models and Severity Levels. (RQ2, \ref{subsec:llm_variants})} The best performance within each category is highlighted in \textbf{bold}, and the overall best performance across all categories is shaded in \textcolor{blue!80}{blue}. 
}
\vspace{-0.1in}
\label{tab:llm-ablation}
\begin{tabular}{cl|ccc|ccc|ccc|ccc|cc}
\toprule
\multirow{2}{*}{} & \multirow{2}{*}{Model} & \multicolumn{3}{c|}{Normal} & \multicolumn{3}{c|}{Mild} & \multicolumn{3}{c|}{Moderate} & \multicolumn{3}{c|}{Severe} & \multirow{2}{*}{Acc} & \multirow{2}{*}{Macro-F1} \\
& & Pre & Rec & F1 & Pre & Rec & F1 & Pre & Rec & F1 & Pre & Rec & F1 & &\\
\midrule
\multirow{7}{*}{\rotatebox{90}{Qwen3-4B}} 
& Zero-shot & 0.5523 & \cellcolor{blue!20}\textbf{1.0000} & \textbf{0.7116} & 0.0000 & 0.0000 & 0.0000 & 0.0000 & 0.0000 & 0.0000 & 0.0000 & 0.0000 & 0.0000 & 0.5523 & 0.1779 \\
& Recency-based & \cellcolor{blue!20}\textbf{0.6976} & 0.6229 & 0.6582 & 0.3941 & 0.4381 & 0.4149 & \textbf{0.1941} & 0.2534 & \textbf{0.2198} & 0.2775 & \textbf{0.2647} & 0.2710 & 0.5165 & 0.3910 \\
& Similarity-based & 0.6944 & 0.7272 & 0.7104 & \textbf{0.4411} & 0.4405 & \textbf{0.4408} & 0.1906 & 0.1667 & 0.1778 & \textbf{0.3438} & 0.2311 & \textbf{0.2764} & \textbf{0.5660} & \textbf{0.4014} \\
& Statistical-pop & 0.5523 & \cellcolor{blue!20}\textbf{1.0000} & \textbf{0.7116} & 0.0000 & 0.0000 & 0.0000 & 0.0000 & 0.0000 & 0.0000 & 0.0000 & 0.0000 & 0.0000 & 0.5523 & 0.1779 \\
& Statistical-ind & 0.6612 & 0.3361 & 0.4457 & 0.3188 & \textbf{0.5217} & 0.3957 & 0.1538 & \textbf{0.3356} & 0.2109 & 0.0247 & 0.0084 & 0.0125 & 0.3798 & 0.2662 \\
& Pattern-based & 0.6574 & 0.4044 & 0.5008 & 0.3313 & 0.5174 & 0.4039 & 0.1213 & 0.2055 & 0.1525 & 0.0982 & 0.0462 & 0.0629 & 0.4069 & 0.2800 \\
& LoRA & 0.5491 & 0.9646 & 0.6999 & 0.3333 & 0.0018 & 0.0036 & 0.0857 & 0.0274 & 0.0415 & 0.0000 & 0.0000 & 0.0000 & 0.0000 & 0.1863 \\
\hline
\multirow{8}{*}{\rotatebox{90}{Qwen3-8B}}
& Zero-shot & 0.0000 & 0.0000 & 0.0000 & 0.0000 & 0.0000 & 0.0000 & 0.0867 & \cellcolor{blue!20}\textbf{0.9452} & 0.1588 & 0.0530 & 0.0882 & 0.0662 & 0.0841 & 0.0563 \\
& Recency-based & \textbf{0.6924} & 0.8183 & \textbf{0.7501} & 0.4881 & 0.3490 & 0.4070 & \cellcolor{blue!20}\textbf{0.2168} & 0.1941 & \textbf{0.2048} & 0.3879 & 0.3782 & 0.3830 & 0.0000 & 0.4362 \\
& Similarity-based & 0.6920 & 0.8151 & 0.7486 & 0.4833 & 0.3527 & \textbf{0.4078} & 0.2078 & 0.1941 & 0.2007 & 0.4455 & 0.3782 & \cellcolor{blue!20}\textbf{0.4091} & 0.0000 & \cellcolor{blue!20}\textbf{0.4415} \\
& Statistical-pop & 0.5896 & 0.0357 & 0.0673 & 0.2583 & \textbf{0.3813} & 0.3080 & 0.0775 & 0.2352 & 0.1166 & 0.0681 & 0.3571 & 0.1143 & 0.1769 & 0.1516 \\
& Statistical-ind & 0.6822 & 0.1488 & 0.2443 & 0.1530 & 0.0988 & 0.1201 & 0.0852 & 0.4612 & 0.1438 & 0.0769 & 0.3613 & 0.1268 & 0.1692 & 0.1588 \\
& Pattern-based & 0.6237 & 0.3610 & 0.4573 & 0.2969 & 0.2373 & 0.2638 & 0.1123 & 0.2763 & 0.1597 & 0.0875 & \textbf{0.4160} & 0.1446 & 0.3172 & 0.2564 \\
& LoRA & 0.6413 & \textbf{0.8690} & 0.7380 & \textbf{0.4945} & 0.2721 & 0.3510 & 0.2090 & 0.1484 & 0.1736 & \cellcolor{blue!20}\textbf{0.4773} & 0.1765 & 0.2577 & \textbf{0.5869} & 0.3801 \\
& PromptTuning & 0.5668 & 0.8522 & 0.6808 & 0.3531 & 0.1708 & 0.2303 & 0.1974 & 0.0342 & 0.0584 & 0.2500 & 0.0042 & 0.0083 & 0.5279 & 0.2444 \\
\hline
\multirow{7}{*}{\rotatebox{90}{Qwen3-14B}}
& Zero-shot & 0.5523 & \cellcolor{blue!20}\textbf{1.0000} & 0.7116 & 0.0000 & 0.0000 & 0.0000 & 0.0000 & 0.0000 & 0.0000 & 0.0000 & 0.0000 & 0.0000 & 0.5523 & 0.1779 \\
& Recency-based & \textbf{0.6941} & 0.8134 & 0.7490 & 0.4883 & 0.3423 & 0.4024 & 0.1895 & 0.1895 & 0.1895 & 0.3924 & 0.3908 & 0.3916 & 0.5918 & 0.4331 \\
& Similarity-based & 0.6921 & 0.8179 & \textbf{0.7498} & 0.4876 & \textbf{0.3484} & \textbf{0.4064} & \textbf{0.2039} & 0.1918 & 0.1976 & 0.4272 & 0.3824 & \textbf{0.4035} & \textbf{0.5960} & \textbf{0.4393} \\
& Statistical-pop & 0.6598 & 0.2710 & 0.3842 & 0.1726 & 0.0940 & 0.1217 & 0.0823 & 0.3037 & 0.1294 & 0.0866 & \textbf{0.5420} & 0.1494 & 0.2301 & 0.1962 \\
& Statistical-ind & 0.6442 & 0.7062 & 0.6738 & 0.3663 & 0.2691 & 0.3102 & 0.1480 & 0.1872 & 0.1653 & 0.0816 & 0.0966 & 0.0885 & 0.0000 & 0.3095 \\
& Pattern-based & 0.6521 & 0.5434 & 0.5928 & 0.3106 & 0.2862 & 0.2979 & 0.1730 & \textbf{0.4247} & \cellcolor{blue!20}\textbf{0.2459} & 0.1602 & 0.1387 & 0.1486 & 0.4332 & 0.3213 \\
& LoRA & 0.6189 & 0.9058 & 0.7354 & \cellcolor{blue!20}\textbf{0.5094} & 0.2154 & 0.3027 & 0.1677 & 0.0594 & 0.0877 & \textbf{0.4762} & 0.2521 & 0.3297 & 0.5852 & 0.3639 \\
\hline
\multirow{6}{*}{\rotatebox{90}{GPT-4.1}}
& Zero-shot & 0.0000 & 0.0000 & 0.0000 & 0.3179 & \cellcolor{blue!20}\textbf{0.9780} & \cellcolor{blue!20}\textbf{0.4799} & 0.0930 & 0.0274 & 0.0423 & 0.0000 & 0.0000 & 0.0000 & 0.0000 & 0.1305 \\
& Recency-based & 0.6905 & \textbf{0.8204} & \cellcolor{blue!20}\textbf{0.7499} & \textbf{0.4889} & 0.3490 & 0.4073 & \textbf{0.2050} & 0.1872 & \textbf{0.1957} & \textbf{0.4375} & 0.3824 & \textbf{0.4081} & \cellcolor{blue!20}\textbf{0.5972} & \textbf{0.4402} \\
& Similarity-based & \textbf{0.6938} & 0.7735 & 0.7315 & 0.4532 & 0.2923 & 0.3553 & 0.1997 & 0.2877 & 0.2357 & 0.3278 & 0.4118 & 0.3650 & 0.5631 & 0.4219 \\
& Statistical-pop & 0.8276 & 0.0420 & 0.0800 & 0.2111 & 0.1666 & 0.1862 & 0.0524 & 0.1826 & 0.0814 & 0.0839 & \cellcolor{blue!20}\textbf{0.7773} & 0.1514 & 0.1272 & 0.1248 \\
& Statistical-ind & 0.7750 & 0.2980 & 0.4305 & 0.2428 & 0.1586 & 0.1919 & 0.0964 & 0.2945 & 0.1453 & 0.1016 & 0.7101 & 0.1777 & 0.2725 & 0.2363 \\
& Pattern-based & 0.6591 & 0.5504 & 0.5999 & 0.2538 & 0.1208 & 0.1637 & 0.1248 & \textbf{0.3333} & 0.1816 & 0.0957 & 0.3361 & 0.1490 & 0.3860 & 0.2735 \\
\bottomrule
\end{tabular}
\end{table}

(1) \textit{\textbf{The scaling law phenomenon does not consistently hold for the mental health forecasting task}}. Specifically, within the Qwen3 model family, increasing model size from 4B to 14B parameters does not yield a clear or consistent improvement in overall Macro-F1 performance. For example, under the Similarity-based setting, the Macro-F1 scores are 0.4014, 0.4415, and 0.4393 for the 4B, 8B, and 14B models, respectively. Similarly, the LoRA-based PEFT results in 0.1863, 0.3801, and 0.3639. These results indicate that simply scaling up model parameters offers diminishing returns, suggesting that performance has plateaued and that larger models may not inherently confer greater advantages for this domain-specific forecasting task.

(2) \textit{\textbf{Regarding different adaptation strategies, PEFT always performs better than Zero-shot but worse than In-context Learning}}. In particular, the Similarity-based and Recency-based strategies achieve the highest Macro-F1 scores (exceeding 0.4) across all models. This demonstrates that providing the model with relevant contextual examples---either those most similar to the target sample or most recent in temporal proximity---significantly enhances its inference accuracy and adaptation to the forecasting task.

(3) \textit{\textbf{Among all In-Context Learning strategies, the Similarity-based and Recency-based methods, both of which select the most similar or most recent sample as the example in few-shot learning, consistently achieve the best performance}}. This suggests that providing LLMs with the most contextually relevant examples yields the greatest improvement in inference accuracy. \textit{\textbf{The Pattern-based strategy, which leverages high-level behavioral patterns derived from stronger LLMs as contextual knowledge, outperforms the Statistical-based methods that use aggregated numerical summaries (mean and variance) of historical data}}. This highlights that semantic representations are more effective than purely numerical statistics for guiding model reasoning.
Finally, between the two Statistical-based variants, \textit{\textbf{Statistical-individual (which uses each user’s own feature statistics) surpasses Statistical-population (which uses global population statistics)}}, underscoring the importance of personalization in mental health forecasting. This is also consistent with our analysis in Section~\ref{subsub:individual-analysis}. 

(4) \textit{\textbf{In terms of the PEFT strategies, LoRA consistently outperforms Prompt Tuning}} for the mental health forecasting task. The possible explanation is that LoRA introduces trainable low-rank matrices directly into the attention layers of the model, enabling more profound adaptation of internal representations to domain-specific behavioral data. In contrast, Prompt Tuning only adjusts input-level embeddings, which primarily influence how instructions are interpreted rather than how representations are computed. This shallow modification may be insufficient for complex, domain-specific reasoning required in mental health forecasting.

(5) \textit{\textbf{With respect to the different mental health classes, approximately 80\% of the reported results achieve high F1-scores for the Normal class}}. However, under the Zero-shot setting, Qwen3 tends to classify nearly all samples as Normal, resulting in high Recall but low Precision. This suggests that, in the absence of reference examples, Qwen3 overpredicts the Normal class, whereas GPT-4.1 exhibits a bias toward predicting Mild, reflecting differences in the prior knowledge encoded during pretraining.
\textit{\textbf{For the Mild, Moderate, and Severe classes, the performance of all models drops markedly compared to Normal}}. Consistent with the intra-/inter-class similarity analysis in Section~\ref{subsubsec:analysis_population}, the behavioral patterns associated with these three abnormal states are highly overlapping, particularly between Mild and Moderate/Severe, making them inherently difficult to distinguish. Nevertheless, the Similarity-/Recency-based strategies continue to achieve the best overall performance, demonstrating their superior ability to capture subtle behavioral differences between adjacent mental health states.

\subsection{Comparison Across Different Modeling Approaches on Early Prediction Capability (RQ3)}\label{subsec:early}

To assess the temporal sensitivity of different modeling approaches, we design an incremental evaluation procedure that simulates a progressively expanding observation window. For each user, the model is first provided with only the first week of data to forecast the mental health status at the end of the second week. The observation window is then extended one day at a time, with the model continuously forecasting the same target (the end of the second week). This process is equivalent to gradually decreasing the value of $\Delta$ and increasing the value of $T$ in Equation~\eqref{eq:forecast}. After each increment, model performance is re-evaluated to track how predictive accuracy evolves as additional behavioral data becomes available.
If the overall accuracy does not increase significantly with more data, this suggests that the model possesses strong early predictive capability: a particularly valuable property in mental health forecasting, where early detection enables timely interventions to mitigate potential risks.
We select the top two best-performing methods from each category of modeling approaches: Logistic Regression (LR) and SVM for Traditional ML models; Temporal Convolutional Networks (TCN) and Transformer (TSF) for DL models; and Qwen3-8B and GPT-4.1 for LLMs~\footnote{In the following experiments, for each model, we use its best-performing configuration from Table~\ref{tab:overall} as the base model. Building on this foundation, we then investigate the effects of specific strategies by systematically varying them and analyzing their impact on performance.}.

\begin{figure}[t]
	\centering
    \begin{subfigure}{0.45\textwidth}
    \centering
	\begin{tikzpicture}
        \begin{axis}[
            width=1\textwidth,
            height=0.7\textwidth,
            xlabel={$T$ (Days)},
            ylabel={Accuracy},
            symbolic x coords={7,8,9,10,11,12,13,14},
            xtick={7,8,9,10,11,12,13,14},
            ymin=0.58, ymax=0.70,
            ytick={0.58,0.58,0.60,0.62,0.64,0.66},
            ylabel style={font=\footnotesize},
            xlabel style={font=\footnotesize},
            tick label style={font=\footnotesize},
            legend style={at={(0.5,0.98)}, font=\scriptsize, anchor=north, legend columns=3, draw=none},
            ymajorgrids=true,
            grid style=dashed,
            enlarge x limits=0.08,
            scaled ticks=false
        ]
        \addplot[color=blue, mark=triangle, line width=1pt, smooth] coordinates {
            (7,0.6424) 
            (8,0.6418) 
            (9,0.6418) 
            (10,0.6418)
            (11,0.6430) 
            (12,0.6432) 
            (13,0.6434) 
            (14,0.6442)
            };
        \addplot[color=red, mark=triangle, line width=1pt, smooth] coordinates {
            (7,0.6538) (8,0.6521) (9,0.6577) (10,0.6587)
    (11,0.6606) (12,0.6594) (13,0.6606) (14,0.6602)
        };
        \addplot[color=green!70!black, mark=triangle, line width=1pt, smooth] coordinates {
            (7,0.6515) (8,0.6521) (9,0.6504) (10,0.6504)
    (11,0.6513) (12,0.6494) (13,0.6529) (14,0.6527)
        };
        \addplot[color=orange, mark=triangle, line width=1pt, smooth] coordinates {
            (7,0.6417) (8,0.6409) (9,0.6415) (10,0.6407)
    (11,0.6395) (12,0.6411) (13,0.6393) (14,0.6413)
        };
        \addplot[color=purple, mark=triangle, line width=1pt, smooth] coordinates {
            (7,0.5958) (8,0.5964) (9,0.5889) (10,0.5904)
    (11,0.5943) (12,0.5949) (13,0.5899) (14,0.5955)
        };
        \addplot[color=black, mark=triangle, line width=1pt, smooth] coordinates {
            (7,0.5970) (8,0.5966) (9,0.5878) (10,0.5961)
    (11,0.5894) (12,0.5991) (13,0.5977) (14,0.5981)
        };
        \legend{LR, SVM, TCN, TSF, Qwen3-8B, GPT-4.1}
        \end{axis}
        \end{tikzpicture}
        \vspace{-0.1in}
        \caption{\footnotesize{All classes}}
    \end{subfigure}
    \vspace{0.1in}
    \hspace{0.2in}
	\begin{subfigure}{0.45\textwidth}
    \centering
	\begin{tikzpicture}
        \begin{axis}[
            width=1\textwidth,
            height=0.7\textwidth,
            xlabel={$T$ (Days)},
            ylabel={Macro-F1},
            symbolic x coords={7,8,9,10,11,12,13,14},
            xtick={7,8,9,10,11,12,13,14},
            ymin=0.42, ymax=0.65,
            ytick={0.42,0.46,0.50,0.54,0.58},
            ylabel style={font=\footnotesize},
            xlabel style={font=\footnotesize},
            tick label style={font=\footnotesize},
            legend style={at={(0.5,0.98)}, font=\scriptsize, anchor=north, legend columns=3, draw=none},
            ymajorgrids=true,
            grid style=dashed,
            enlarge x limits=0.08,
            scaled ticks=false
        ]
        \addplot[color=blue, mark=triangle, line width=1pt, smooth] coordinates {
    (7,0.5086) (8,0.5088) (9,0.5094) (10,0.5093)
    (11,0.5103) (12,0.5112) (13,0.5107) (14,0.5118)
}; 
\addplot[color=red, mark=triangle, line width=1pt, smooth] coordinates {
    (7,0.4794) (8,0.4769) (9,0.4820) (10,0.4837)
    (11,0.4818) (12,0.4800) (13,0.4829) (14,0.4810)
}; 
\addplot[color=green!70!black, mark=triangle, line width=1pt, smooth] coordinates {
    (7,0.5456) (8,0.5457) (9,0.5431) (10,0.5431)
    (11,0.5425) (12,0.5388) (13,0.5419) (14,0.5401)
}; 
\addplot[color=orange, mark=triangle, line width=1pt, smooth] coordinates {
    (7,0.5808) (8,0.5807) (9,0.5811) (10,0.5806)
    (11,0.5800) (12,0.5808) (13,0.5793) (14,0.5808)
}; 
\addplot[color=purple, mark=triangle, line width=1pt, smooth] coordinates {
    (7,0.4415) (8,0.4416) (9,0.4413) (10,0.4414)
    (11,0.4417) (12,0.4426) (13,0.4409) (14,0.4387)
}; 
\addplot[color=black, mark=triangle, line width=1pt, smooth] coordinates {
    (7,0.4411) (8,0.4416) (9,0.4411) (10,0.4412)
    (11,0.4411) (12,0.4420) (13,0.4413) (14,0.4404)
}; 
        \legend{LR, SVM, TCN, TSF, Qwen3-8B, GPT-4.1}
        \end{axis}
        \end{tikzpicture}
        \vspace{-0.1in}
        \caption{\footnotesize{All classes}}
    \end{subfigure}

    \begin{subfigure}{0.45\textwidth}
    \centering
	\begin{tikzpicture}
        \begin{axis}[
            width=1\textwidth,
            height=0.7\textwidth,
            xlabel={$T$ (Days)},
            ylabel={F1},
            symbolic x coords={7,8,9,10,11,12,13,14},
            xtick={7,8,9,10,11,12,13,14},
            ymin=0.74, ymax=0.82,
            ytick={0.74,0.75,0.76,0.77,0.78,0.79},
            ylabel style={font=\footnotesize},
            xlabel style={font=\footnotesize},
            tick label style={font=\footnotesize},
            legend style={at={(0.5,0.98)}, font=\scriptsize, anchor=north, legend columns=3, draw=none},
            ymajorgrids=true,
            grid style=dashed,
            enlarge x limits=0.08,
            scaled ticks=false
        ]
        \addplot[color=blue, mark=triangle, line width=1pt, smooth] coordinates {
            (7,0.7822) (8,0.7816) (9,0.7808) (10,0.7805)
            (11,0.7817) (12,0.7821) (13,0.7819) (14,0.7827)
        }; 
        \addplot[color=red, mark=triangle, line width=1pt, smooth] coordinates {
            (7,0.7893) (8,0.7878) (9,0.7914) (10,0.7927)
            (11,0.7938) (12,0.7930) (13,0.7935) (14,0.7931)
        }; 
        \addplot[color=green!70!black, mark=triangle, line width=1pt, smooth] coordinates {
            (7,0.7829) (8,0.7838) (9,0.7831) (10,0.7827)
            (11,0.7832) (12,0.7809) (13,0.7832) (14,0.7837)
        }; 
        \addplot[color=orange, mark=triangle, line width=1pt, smooth] coordinates {
            (7,0.7746) (8,0.7736) (9,0.7746) (10,0.7735)
            (11,0.7724) (12,0.7739) (13,0.7728) (14,0.7742)
        }; 
        \addplot[color=purple, mark=triangle, line width=1pt, smooth] coordinates {
            (7,0.7486) (8,0.7489) (9,0.7482) (10,0.7495)
            (11,0.7479) (12,0.7491) (13,0.7473) (14,0.7498)
        }; 
        \addplot[color=black, mark=triangle, line width=1pt, smooth] coordinates {
            (7,0.7514) (8,0.7526) (9,0.7517) (10,0.7511)
            (11,0.7522) (12,0.7519) (13,0.7508) (14,0.7501)
        }; 
        \legend{LR, SVM, TCN, TSF, Qwen3-8B, GPT-4.1}
        \end{axis}
        \end{tikzpicture}
        \vspace{-0.1in}
        \caption{\footnotesize{Normal class}}
    \end{subfigure}
    \vspace{0.1in}
    \hspace{0.2in}
	\begin{subfigure}{0.45\textwidth}
    \centering
	\begin{tikzpicture}
        \begin{axis}[
            width=1\textwidth,
            height=0.7\textwidth,
            xlabel={$T$ (Days)},
            ylabel={F1},
            symbolic x coords={7,8,9,10,11,12,13,14},
            xtick={7,8,9,10,11,12,13,14},
            ymin=0.38, ymax=0.59,
            ytick={0.38,0.41,0.44,0.47,0.50,0.53},
            ylabel style={font=\footnotesize},
            xlabel style={font=\footnotesize},
            tick label style={font=\footnotesize},
            legend style={at={(0.5,0.98)}, font=\scriptsize, anchor=north, legend columns=3, draw=none},
            ymajorgrids=true,
            grid style=dashed,
            enlarge x limits=0.08,
            scaled ticks=false
        ]
        \addplot[color=blue, mark=triangle, line width=1pt, smooth] coordinates {
            (7,0.3859) (8,0.3859) (9,0.3876) (10,0.3899)
            (11,0.3907) (12,0.3910) (13,0.3915) (14,0.3919)
        }; 
        
        \addplot[color=red, mark=triangle, line width=1pt, smooth] coordinates {
            (7,0.4907) (8,0.4901) (9,0.4995) (10,0.5000)
            (11,0.5057) (12,0.5045) (13,0.5044) (14,0.5034)
        }; 
        
        \addplot[color=green!70!black, mark=triangle, line width=1pt, smooth] coordinates {
            (7,0.4770) (8,0.4769) (9,0.4707) (10,0.4720)
            (11,0.4759) (12,0.4746) (13,0.4791) (14,0.4768)
        }; 
        
        \addplot[color=orange, mark=triangle, line width=1pt, smooth] coordinates {
            (7,0.5206) (8,0.5199) (9,0.5206) (10,0.5192)
            (11,0.5170) (12,0.5198) (13,0.5173) (14,0.5197)
        }; 
        
        \addplot[color=purple, mark=triangle, line width=1pt, smooth] coordinates {
            (7,0.4078) (8,0.4075) (9,0.4088) (10,0.4082)
            (11,0.4072) (12,0.4085) (13,0.4091) (14,0.4065)
        }; 
        
        \addplot[color=black, mark=triangle, line width=1pt, smooth] coordinates {
            (7,0.3994) (8,0.4003) (9,0.3991) (10,0.3988)
            (11,0.3997) (12,0.4001) (13,0.3982) (14,0.4006)
        }; 
        \legend{LR, SVM, TCN, TSF, Qwen3-8B, GPT-4.1}
        \end{axis}
        \end{tikzpicture}
        \vspace{-0.1in}
        \caption{\footnotesize{Mild class}}
    \end{subfigure}
    \vspace{0.1in}

    \begin{subfigure}{0.45\textwidth}
    \centering
	\begin{tikzpicture}
        \begin{axis}[
            width=1\textwidth,
            height=0.7\textwidth,
            xlabel={$T$ (Days)},
            ylabel={F1},
            symbolic x coords={7,8,9,10,11,12,13,14},
            xtick={7,8,9,10,11,12,13,14},
            ymin=0.15, ymax=0.49,
            ytick={0.15,0.20,0.25,0.30,0.35,0.40},
            ylabel style={font=\footnotesize},
            xlabel style={font=\footnotesize},
            tick label style={font=\footnotesize},
            legend style={at={(0.5,0.98)}, font=\scriptsize, anchor=north, legend columns=3, draw=none},
            ymajorgrids=true,
            grid style=dashed,
            enlarge x limits=0.08,
            scaled ticks=false
        ]
            \addplot[color=blue, mark=triangle, line width=1pt, smooth] coordinates {
             (7,0.2462) (8,0.2435) (9,0.2412) (10,0.2389) (11,0.2446) (12,0.2439) (13,0.2416) (14,0.2446)
            }; 
            
            \addplot[color=red, mark=triangle, line width=1pt, smooth] coordinates {
             (7,0.1637) (8,0.1631) (9,0.1659) (10,0.1686) (11,0.1739) (12,0.1742) (13,0.1806) (14,0.1818)
            }; 
            
            \addplot[color=green!70!black, mark=triangle, line width=1pt, smooth] coordinates {
             (7,0.3437) (8,0.3457) (9,0.3521) (10,0.3534) (11,0.3515) (12,0.3521) (13,0.3621) (14,0.363)
            }; 
            
            \addplot[color=orange, mark=triangle, line width=1pt, smooth] coordinates {
             (7,0.3868) (8,0.388) (9,0.3881) (10,0.3888) (11,0.3897) (12,0.3884) (13,0.3862) (14,0.3884)
            }; 
            
            \addplot[color=purple, mark=triangle, line width=1pt, smooth] coordinates {
             (7,0.2007) (8,0.1995) (9,0.2004) (10,0.2015) (11,0.1998) (12,0.2018) (13,0.1989) (14,0.1889)
            }; 
            
            \addplot[color=black, mark=triangle, line width=1pt, smooth] coordinates {
             (7,0.2063) (8,0.2066) (9,0.2055) (10,0.2071) (11,0.2058) (12,0.2068) (13,0.2075) (14,0.2051)
            }; 
        \legend{LR, SVM, TCN, TSF, Qwen3-8B, GPT-4.1}
        \end{axis}
        \end{tikzpicture}
        \vspace{-0.1in}
        \caption{\footnotesize{Moderate class}}
    \end{subfigure}
    \hspace{0.2in}
	\begin{subfigure}{0.45\textwidth}
    \centering
	\begin{tikzpicture}
        \begin{axis}[
            width=1\textwidth,
            height=0.7\textwidth,
            xlabel={$T$ (Days)},
            ylabel={F1},
            symbolic x coords={7,8,9,10,11,12,13,14},
            xtick={7,8,9,10,11,12,13,14},
            ymin=0.395, ymax=0.75,
            ytick={0.40,0.45,0.50,0.55,0.60,0.65},
            ylabel style={font=\footnotesize},
            xlabel style={font=\footnotesize},
            tick label style={font=\footnotesize},
            legend style={at={(0.5,0.98)}, font=\scriptsize, anchor=north, legend columns=3, draw=none},
            ymajorgrids=true,
            grid style=dashed,
            enlarge x limits=0.08,
            scaled ticks=false
        ]
            \addplot[color=blue, mark=triangle, line width=1pt, smooth] coordinates {
             (7,0.6201) (8,0.6240) (9,0.6278) (10,0.6278) 
             (11,0.6240) (12,0.6278) (13,0.6278) (14,0.6278)
            }; 
            
            \addplot[color=red, mark=triangle, line width=1pt, smooth] coordinates {
             (7,0.4739) (8,0.4667) (9,0.4712) (10,0.4734)
             (11,0.4537) (12,0.4483) (13,0.4532) (14,0.4455)
            }; 
            
            \addplot[color=green!70!black, mark=triangle, line width=1pt, smooth] coordinates {
             (7,0.5789) (8,0.5762) (9,0.5666) (10,0.5645)
             (11,0.5594) (12,0.5477) (13,0.5431) (14,0.5371)
            }; 
            
            \addplot[color=orange, mark=triangle, line width=1pt, smooth] coordinates {
             (7,0.6411) (8,0.6411) (9,0.6411) (10,0.6411)
             (11,0.6411) (12,0.6411) (13,0.6411) (14,0.6411)
            }; 
            
            \addplot[color=purple, mark=triangle, line width=1pt, smooth] coordinates {
             (7,0.4091) (8,0.4105) (9,0.4078) (10,0.4063)
             (11,0.4118) (12,0.4108) (13,0.4082) (14,0.4095)
            }; 
            
            \addplot[color=black, mark=triangle, line width=1pt, smooth] coordinates {
             (7,0.4072) (8,0.4068) (9,0.4082) (10,0.4079)
             (11,0.4065) (12,0.4091) (13,0.4085) (14,0.4058)
            }; 
        \legend{LR, SVM, TCN, TSF, Qwen3-8B, GPT-4.1}
        \end{axis}
        \end{tikzpicture}
        \vspace{-0.1in}
        \caption{\footnotesize{Severe class}}
    \end{subfigure}
 	\vspace{-0.1in}
	\caption{\textbf{Comparison of different models on early prediction capability (RQ3, \ref{subsec:early}).} 
	}\label{fig:day_class_acc}
 	\vspace{-0.1in}
\end{figure}
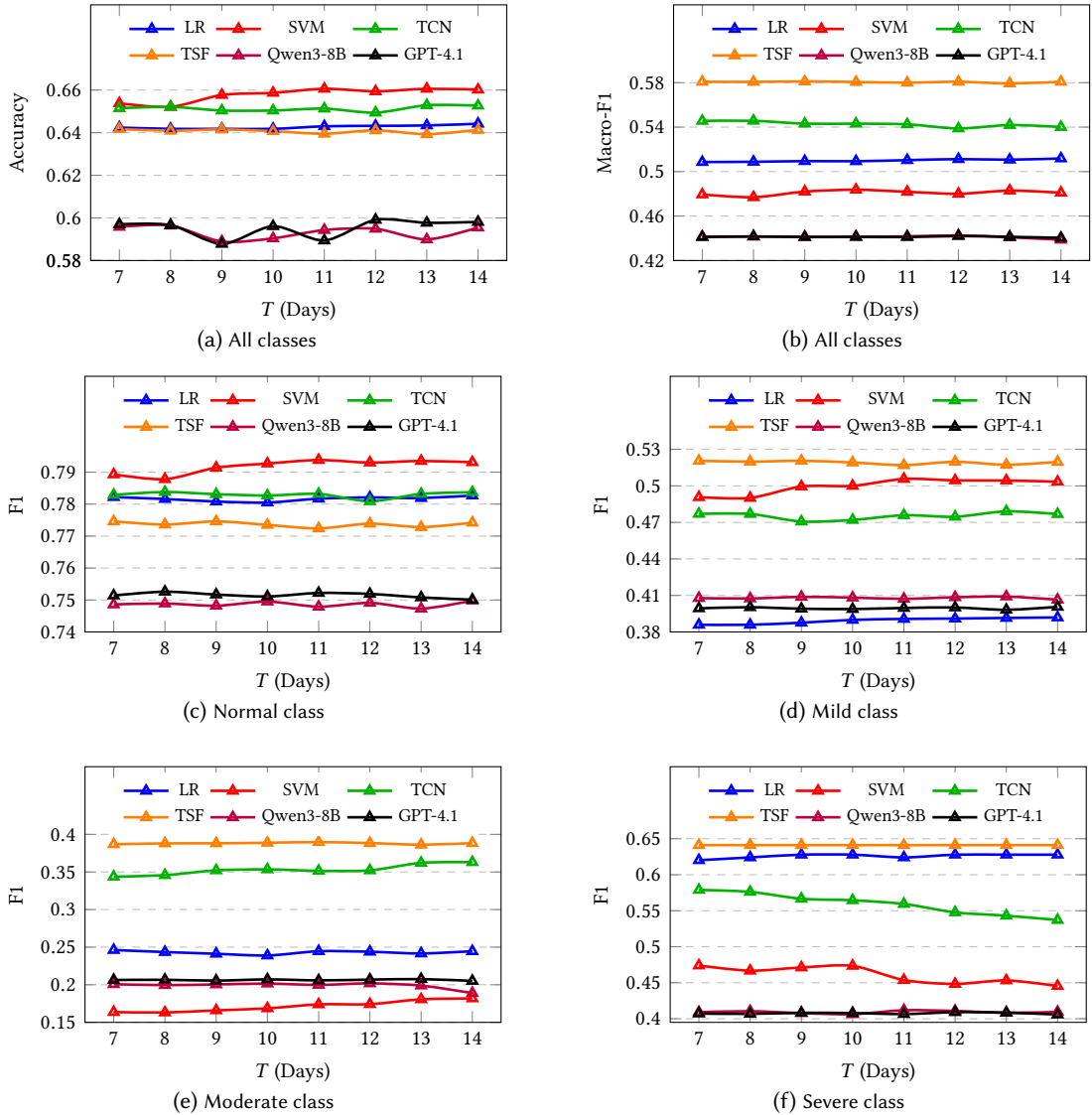

Fig.~\ref{fig:day_class_acc}(a–b) present the overall Accuracy and Macro-F1 across all classes, while Fig.~\ref{fig:day_class_acc}(c–f) illustrate the F1-scores for each individual class. Several key observations can be drawn from these results.
First, \textit{\textbf{in terms of overall performance}}, the accuracy of traditional ML models (LR and SVM) gradually increases as the observation window $T$ expands. In contrast, the DL models and LLMs achieve their best performance at the earliest stage ($T=7$), indicating that \textit{\textbf{DL models and LLMs possess stronger early prediction capability compared to traditional ML approaches}}. However, for LLMs, we observe noticeable performance fluctuations when $T\in[9,13]$, suggesting that additional data may introduce noise or irrelevant information, thereby hindering forecasting stability. Moreover, unlike accuracy, the variations in Macro-F1 across different $T$ values are relatively minor, implying that the class-level balance of predictions remains largely consistent over time. 
Second, \textit{\textbf{regarding the performance across individual classes, similar overall trends can be observed}}. The main differences are twofold: (1) the performance of LLMs remains relatively stable across varying values of $T$, especially on the Mild and Severe classes; and (2) the performance of SVM and TCN gradually declines as $T$ increases for the Severe class, suggesting that they possess stronger early prediction capability for identifying severe mental health conditions at earlier stages.  

\subsection{Comparison Across Different Feature Representation Strategies (RQ4)}\label{subsec:feature_rep_results}

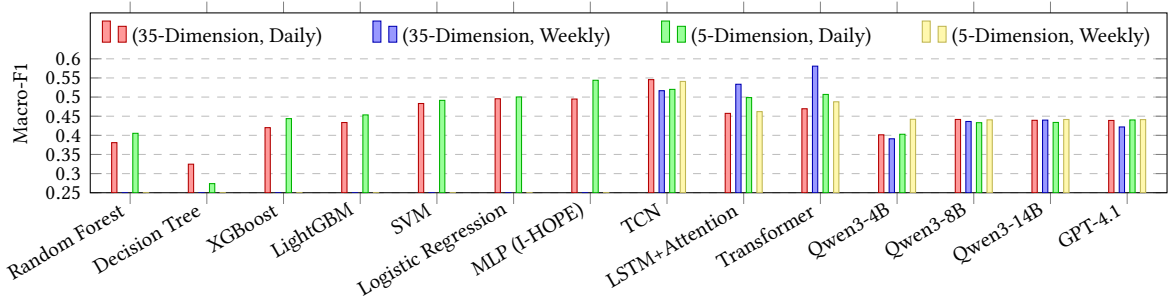
\begin{figure}[t]
	\centering
	\begin{tikzpicture}
	\begin{axis}[
    ybar, 
    width=1\linewidth,
    height=0.25\linewidth,
    bar width=2pt,
    enlarge x limits= 0.04, 
    symbolic x  coords = {1,2,3,4,5,6,7,8,9,10,11,12,13,14},
    xtick = {1,2,3,4,5,6,7,8,9,10,11,12,13,14},
    xticklabels={Random Forest, Decision Tree, XGBoost, LightGBM, SVM, Logistic Regression, MLP (I-HOPE), TCN, LSTM+Attention, Transformer, Qwen3-4B, Qwen3-8B, Qwen3-14B, GPT-4.1},
    xticklabel style={rotate=30,anchor=east,font = \footnotesize},
    ymin=0.25, ymax=0.72, 
    ylabel = {Macro-F1},
    ytick={0.25,0.30,0.35,0.40,0.45,0.50,0.55,0.60},
    ylabel style ={font = \footnotesize},
	tick label style={font = \footnotesize},
    yticklabel style = {/pgf/number format/.cd,fixed,precision=3},
    ymajorgrids=true,
    grid style=dashed,
	legend style={at={(0.5,0.99)},font = \footnotesize, anchor=north,legend columns=4,/tikz/every even column/.style={column sep=0.6cm},legend cell align={left},draw=none},
	ymajorgrids=true,
	grid style=dashed,
	]
	\addplot[color={red!70!black}, fill={red!40!white}] coordinates {
		(1, 0.3809)
    	(2, 0.3246)
    	(3, 0.4201)
    	(4, 0.4335)
        (5,0.4832)
        (6,0.4956)
        (7,0.4949)
        (8,0.5456)
        (9,0.4574)
        (10,0.4695)
        (11,0.4014)
        (12,0.4415)
        (13,0.4393)
        (14,0.4389)
	}; 
	\addplot[color={blue!70!black}, fill={blue!40!white}] coordinates {
        (1,0)      
        (2,0)      
        (3,0)      
        (4,0)      
        (5,0)      
        (6,0)      
        (7,0)      
        (8,0.5167) 
        (9,0.5335) 
        (10,0.5808) 
        (11,0.3910) 
        (12,0.4362) 
        (13,0.4398) 
        (14,0.4219) 
    }; 
    \addplot[color={green!70!black}, fill={green!40!white}] coordinates {
        (1,0.4053)
        (2,0.2738)
        (3,0.4437)
        (4,0.4534)
        (5,0.4914)
        (6,0.5003)
        (7,0.5439)
        (8,0.5201)
        (9,0.4987)
        (10,0.5069)
        (11,0.4028)
        (12,0.4331)
        (13,0.4337)
        (14,0.4402)
    }; 
	\addplot[color={yellow!70!black}, fill={yellow!40!white}] coordinates {
        (1,0)
        (2,0)
        (3,0)
        (4,0)
        (5,0)
        (6,0)
        (7,0)
        (8,0.5409)
        (9,0.4620)
        (10,0.4877)
        (11,0.4419)
        (12,0.4405)
        (13,0.4414)
        (14,0.4411)
    }; 
	\legend{(35-Dimension, Daily), (35-Dimension, Weekly), (5-Dimension, Daily), (5-Dimension, Weekly)}
	\end{axis}
	\end{tikzpicture}
    \vspace{-0.1in}
	\caption{\textbf{Impact of Feature Granularity on Model Performance in Macro-F1 (RQ4, \ref{subsec:feature_rep_results}).}
	}\label{fig:feature_f1}
 \vspace{-0.1in}
\end{figure}

\begin{figure}[t]
	\centering
	\begin{tikzpicture}
	\begin{axis}[
    ybar, 
    width=1\linewidth,
    height=0.25\linewidth,
    bar width=2pt,
    enlarge x limits= 0.04, 
    symbolic x  coords = {1,2,3,4,5,6,7,8,9,10,11,12,13,14},
    xtick = {1,2,3,4,5,6,7,8,9,10,11,12,13,14},
    xticklabels={Random Forest, Decision Tree, XGBoost, LightGBM, SVM, Logistic Regression, MLP (I-HOPE), TCN, LSTM+Attention, Transformer, Qwen3-4B, Qwen3-8B, Qwen3-14B, GPT-4.1},
    xticklabel style={rotate=30,anchor=east,font = \footnotesize},
    ymin=0.35, ymax=0.77, 
    ylabel = {Accuracy},
    ytick={0.35,0.40,0.45,0.50,0.55,0.60,0.65},
    ylabel style ={font = \footnotesize},
	tick label style={font = \footnotesize},
    yticklabel style = {/pgf/number format/.cd,fixed,precision=3},
    ymajorgrids=true,
    grid style=dashed,
	legend style={at={(0.5,0.99)},font = \footnotesize, anchor=north,legend columns=4,/tikz/every even column/.style={column sep=0.6cm},legend cell align={left},draw=none},
	ymajorgrids=true,
	grid style=dashed,
	]
	\addplot[color={red!70!black}, fill={red!40!white}] coordinates {
        (1,0.6030)      
        (2,0.5796)      
        (3,0.6312)      
        (4,0.6449)      
        (5,0.6424)      
        (6,0.6086)      
        (7,0.6426)      
        (8,0.6515)      
        (9,0.6480)      
        (10,0.6202)     
        (11,0.5660)     
        (12,0.5958)     
        (13,0.5960)     
        (14,0.5858)     
        }; 
	\addplot[color={blue!70!black}, fill={blue!40!white}] coordinates {
        (1,0)           
        (2,0)           
        (3,0)           
        (4,0)           
        (5,0)           
        (6,0)           
        (7,0)           
        (8,0.65616)     
        (9,0.64416)     
        (10,0.64164)    
        (11,0.51653)    
        (12,0.59640)    
        (13,0.59582)    
        (14,0.56314)    
        }; 
    \addplot[color={green!70!black}, fill={green!40!white}] coordinates {
        (1,0.6190)      
        (2,0.5699)      
        (3,0.6401)      
        (4,0.6436)      
        (5,0.6565)      
        (6,0.6150)      
        (7,0.6662)      
        (8,0.6480)      
        (9,0.6595)      
        (10,0.6378)     
        (11,0.4974)     
        (12,0.5918)     
        (13,0.5950)     
        (14,0.5972)     
        }; 
	\addplot[color={yellow!70!black}, fill={yellow!40!white}] coordinates {
        (1,0)           
        (2,0)           
        (3,0)           
        (4,0)           
        (5,0)           
        (6,0)           
        (7,0)           
        (8,0.6589)      
        (9,0.6565)      
        (10,0.6196)     
        (11,0.5954)     
        (12,0.5974)     
        (13,0.5974)     
        (14,0.5970)     
        }; 
	\legend{(35-Dimension, Daily), (35-Dimension, Weekly), (5-Dimension, Daily), (5-Dimension, Weekly)}
	\end{axis}
	\end{tikzpicture}
    \vspace{-0.1in}
	\caption{\textbf{Impact of Feature Granularity on Model Performance in Accuracy (RQ4, \ref{subsec:feature_rep_results}).}
	}\label{fig:feature_acc}
 \vspace{-0.1in}
\end{figure}
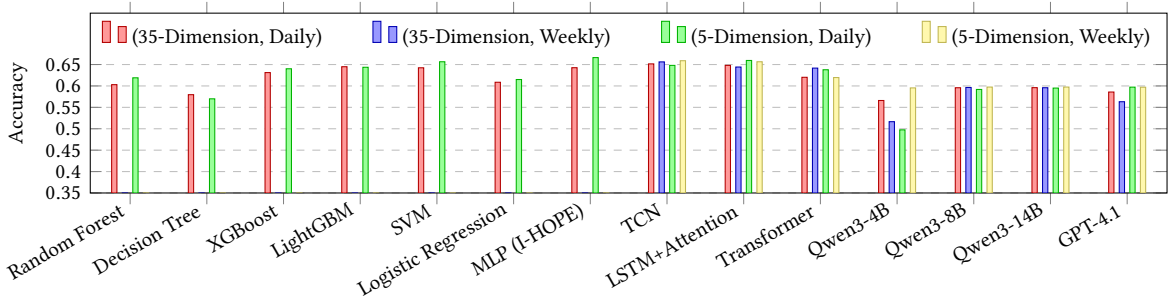
We further examine the impacts of different feature representation strategies on the forecasting performance.
Fig.~\ref{fig:feature_f1} and Fig.~\ref{fig:feature_acc} respectively present the Macro-F1 and Accuracy scores of different models~\footnote{Although Traditional ML and MLP models cannot inherently process sequential data, their inputs can be organized using the Sequential Flattening strategy introduced in Section~\ref{subsec:traditional-ml}. However, we empirically find that this strategy results in poor forecasting performance. Therefore, their inputs are structured using only the Statistical Aggregation strategy, where temporal features are aggregated by averaging across time. Consequently, these models are compared only across feature dimensionalities (35- vs. 5-Dimension) without including weekly-level results. In contrast, models capable of handling sequential data (i.e. TCN, LSTM-Attention, Transformer, and LLMs) report results for both daily and weekly granularities.} under various feature granularities: (35-Dimension, Daily), (35-Dimension, Weekly), (5-Dimension, Daily), and (5-Dimension, Weekly), as introduced in Section~\ref{subsec:feature_representation}.
Several key observations can be drawn from these results.

(1) \textbf{\textit{For traditional ML and MLP, applying 5-Dimension feature achieves better performance than 35-Dimension feature}} (except Decision Tree). This suggests that the 35-Dimension feature may contain noise or redundant information that confused these models. Aggregating the feature set to the five most critical ones acts as a form of feature selection, allowing the models to build more robust and generalizable decision boundaries.

(2) \textit{\textbf{For sequential DL models, LSTM and Transformer exhibit improved performance when either the feature dimension or temporal duration is reduced}}, suggesting these architectures are sensitive to redundancy and noise in high-dimensional or overly fine-grained temporal features. Specifically, directly using the (35-Dimension, Daily) setting yields the worst results, indicating that the finest temporal granularity introduces unnecessary fluctuations and short-term noise that hinder generalization. In contrast, the \textit{\textbf{TCN achieves its best performance under {(35-Dimension, Daily)} configuration}}, implying that its convolutional structure can effectively exploit local temporal dependencies and benefit from higher-frequency information. This indicates CNN-based temporal models are more robust to fine-grained time steps and can capitalize on dense temporal patterns, unlike sequential models that overfit or saturate with excessive granularity.

(3) \textit{\textbf{For LLMs, they consistently achieve their best performance with the (5-Dimension, Weekly) setting}}. This demonstrates that LLMs, when used for this task via prompting, perform best with high-level, concise, and aggregated information. Noisy daily data or an excessive number of features likely makes it harder for the LLM to discern the core signal within the text-based prompt. A weekly summary with only the most important features provides a clean, potent signal that the model can more effectively reason about.

\subsection{Comparison Between Personalization and Non-Personalization (RQ5)}\label{subsec:personalization}

To examine the effect of personalization modeling, we evaluate each model with and without user identifiers as input features. Table~\ref{tab:personalization} shows the performance difference after applying personalization modeling. Positive values indicate improvement with personalization, while negative values (shaded in \textcolor{blue!80}{blue}) mean a decline in performance. 
Note that for the LLMs, we report the results of different backbone models (Qwen3) fine-tuned with PEFT using LoRA, since Zero-shot and In-context Learning strategies cannot effectively leverage user identifier information. For instance, in the Zero-shot setting, simply appending a user identifier provides minimal personalization cues for inference and fails to capture user-specific behavioral patterns. Several observations can be noted. 

First, \textit{\textbf{DL models benefit the most from personalization by a large margin}}. All DL models show a significant increase in Macro-F1 score, with improvements ranging from +0.2894 to a remarkable +0.3635 for MLP. This demonstrates that integrating user embeddings is a highly effective method to achieve personalization into these architectures.
Second, \textit{\textbf{ML models show moderate but consistent benefits}}. The Macro-F1 improvements are positive across the board, ranging from +0.1144 (Decision Tree) to +0.2504 (Logistic Regression). This suggests that user one-hot encoding successfully helps these models capture user-level variations.
Third, \textit{\textbf{the personalization strategy used for LLMs (adding user identifier to the prompt) is largely ineffective and even detrimental}}. The changes in Macro-F1 are minimal and even negative for the largest model, Qwen3-14B (-0.0377). This shows that simply adding a user ID to a prompt is not a meaningful way to achieve personalization, as LLMs cannot encode each user’s complete historical behaviors and labels within a single prompt. Consequently, even with PEFT, the model fails to establish a consistent mapping between training samples and user identities.
Lastly, \textit{\textbf{across various severity levels, we observe a substantial improvement in the Severe class for both ML and DL models.}} The Precision gains are exceptionally high (e.g., +0.9451 for Random Forest, +0.9213 for MLP, +0.8802 for Transformer). This indicates that personalization is key to helping models correctly and confidently identify high-severity cases, which are often the most critical and challenging to predict. 

\begin{table}[t]
\centering
\footnotesize
\setlength{\tabcolsep}{1.5pt}
\caption{\textbf{Impact of Personalization on Model Performance (RQ5, \ref{subsec:personalization}).} Positive values indicate performance improvements with personalization, whereas negative values, shaded in \textcolor{blue!80}{blue}, represent a decline in performance.
}
\vspace{-0.1in}
\begin{tabular}{ll|ccc|ccc|ccc|ccc|cc}
\toprule
\multirow{2}{*}{} & \multirow{2}{*}{Model} & \multicolumn{3}{c|}{Normal} & \multicolumn{3}{c|}{Mild} & \multicolumn{3}{c|}{Moderate} & \multicolumn{3}{c|}{Severe} & \multirow{2}{*}{Acc} & \multirow{2}{*}{Macro-F1} \\
& & Pre & Rec & F1 & Pre & Rec & F1 & Pre & Rec & F1 & Pre & Rec & F1 & \\
\midrule
\multirow{6}{*}{\begin{tabular}[c]{@{}l@{}}ML\end{tabular}}
& Random Forest & 0.0484 & 0.0357 & 0.0473 & 0.1686 & 0.0744 & 0.1055 & \cellcolor{blue!20}-0.1141 & 0.0205 & 0.0361 & 0.9451 & 0.3613 & 0.5228 &0.0617 &0.1779 \\
& Decision Tree & 0.0162 & 0.0403 & 0.0238 & 0.3135 & \cellcolor{blue!20}-0.0244 & \cellcolor{blue!20}-0.0345 & 0.0154 & 0.0342 & 0.0557 & 0.8442 & 0.2731 & 0.4127 &0.0300 & 0.1144 \\
& XGBoost & 0.0713 & 0.0718 & 0.0740 & 0.1658 & 0.0781 & 0.1063 & \cellcolor{blue!20}-0.0107 & 0.0502 & 0.0799 & 0.3077 & 0.3529 & 0.4380 &0.0849 & 0.1746 \\
& LightGBM & 0.0850 & 0.0970 & 0.0912 & 0.1286 & 0.0549 & 0.0800 & 0.0634 & 0.0799 & 0.1015 & 0.3214 & 0.2815 & 0.3236 &0.0907 & 0.1491 \\
& SVM & 0.1322 & 0.1082 & 0.1255 & 0.1714 & 0.1861 & 0.1867 & 0.2147 & 0.0776 & 0.1141 & 0.5697 & 0.3950 & 0.4700 &0.1435 & 0.2241 \\
& Logistic Regression & 0.1535 & 0.4461 & 0.3565 & 0.1705 & 0.0696 & 0.1142 & 0.1648 & 0.1781 & 0.1784 & 0.2719 & 0.4160 & 0.3524 &0.3026 & 0.2504 \\
\hline
\multirow{4}{*}{DL}
& MLP (I-HOPE~\cite{roychowdhury2025predicting}) & 0.1696 & \cellcolor{blue!20}-0.1208 & 0.0798 & 0.2699 & 0.4594 & 0.4914 & 0.3750 & 0.1781 & 0.2415 & 0.9213 & 0.4916 & 0.6411 &0.1166 & 0.3635 \\
& TCN & 0.1295 & 0.2874 & 0.2061 & 0.1788 & 0.2844 & 0.2696 & 0.1787 & 0.1827 & 0.1819 & 0.6278 & 0.2227 & 0.5020 &0.2746 & 0.2899 \\
& LSTM+Attention & 0.1980 & \cellcolor{blue!20}-0.1113 & 0.0789 & 0.1561 & 0.4204 & 0.3300 & 0.1878 & 0.2192 & 0.2458 & 0.7672 & 0.3739 & 0.5028 &0.1076 & 0.2894 \\
& Transformer & 0.2174 & 0.0910 & 0.1524 & 0.1659 & 0.3020 & 0.2466 & 0.0952 & 0.3721 & 0.1981 & 0.8802 & 0.3740 & 0.5802 &0.1947 & 0.2943 \\
\hline
\multirow{3}{*}{LLM}
& Qwen3-4B & 0.0946 & \cellcolor{blue!20}-0.2563 & \cellcolor{blue!20}-0.0572 & \cellcolor{blue!20}-0.0960 & 0.3026 & 0.2027 & 0.0113 & 0.1370 & 0.0775 & \cellcolor{blue!20}-0.1659 & 0.0672 & \cellcolor{blue!20}-0.0023 &\cellcolor{blue!20}-0.0309 & 0.0552 \\
& Qwen3-8B & 0.0424 & \cellcolor{blue!20}-0.0480 & 0.0112 & \cellcolor{blue!20}-0.0077 & 0.0824 & 0.0670 & \cellcolor{blue!20}-0.0202 & 0.0434 & 0.0379 & \cellcolor{blue!20}-0.0193 & 0.1008 & 0.0655 &0.0079 & 0.0454 \\
& Qwen3-14B & \cellcolor{blue!20}-0.0213 & 0.0116 & \cellcolor{blue!20}-0.0088 & 0.0028 & 0.0122 & 0.0090 & 0.0449 & \cellcolor{blue!20}-0.0297 & \cellcolor{blue!20}-0.0199 & 0.0418 & \cellcolor{blue!20}-0.1555 & \cellcolor{blue!20}-0.1312 &0.0006 & \cellcolor{blue!20}-0.0377 \\
\bottomrule
\end{tabular}
\label{tab:personalization}
\end{table}

\subsection{Comparison Across Different Loss Functions for Class Imbalance (RQ6)}\label{subsec:class_imbalance}

We now examine whether different loss functions (Weighted Cross-Entropy Loss and Focal Loss) can help mitigate the class imbalance issue introduced in Section~\ref{subsec:med_class_imbalance}. It is important to note that neither of these loss functions can be directly applied to traditional ML methods, particularly Focal Loss, which is designed for differentiable models optimized through gradient-based learning. For LLMs, the HuggingFace PEFT framework computes token-level cross-entropy internally and does not natively support loss re-weighting without redefining the model’s forward function. Consequently, we apply these loss functions only to the DL models.

Table~\ref{tab:imbalance} presents the performance of four DL models trained with different loss functions. The Macro-F1 scores for the four methods are highlighted in \textcolor{blue}{blue}, \textcolor{red}{red}, \textcolor{green}{green}, \textcolor{orange}{orange}, with darker shades indicating higher values. We emphasize Macro-F1 because it measures a model’s overall performance across all classes, treating each class equally regardless of its sample size. This makes it a fair and balanced metric for model comparison, especially under our imbalanced data distributions, where accuracy alone may be misleading.
First, Focal Loss achieves the highest Macro-F1 for both Transformer (0.5808) and TCN (0.5456), demonstrating its effectiveness in emphasizing hard-to-classify or minority samples.
Second, the simpler DL model (MLP-based I-HOPE) attains its best performance with Weighted Loss (Macro-F1 = 0.5465), which improves recall across most imbalanced classes but often reduces precision, resulting in only modest gains compared to Cross-Entropy Loss (0.5439). This suggests that explicit class weighting can mitigate imbalance but may also distort decision boundaries in more complex models. In contrast, Focal Loss dynamically scales gradients, enabling high-capacity models to learn more balanced representations without overfitting to the majority classes.
Third, LSTM performs most stably with Vanilla Cross Entropy Loss (0.5147), indicating that reweighting mechanisms may introduce optimization instability in recurrent architectures.
Overall, \textit{\textbf{Focal Loss is most effective for models with high-capacity (e.g., Transformer), whereas simpler or sequence-based models (e.g., MLP and LSTM) tend to benefit more from Vanilla or Weighted Cross Entropy losses}}.

\begin{table}[t]
\centering
\footnotesize
\setlength{\tabcolsep}{2pt}
\caption{\textbf{Impact of Different Loss Functions on DL Models under Class Imbalance (RQ6, \ref{subsec:class_imbalance}).} The Macro-F1 for the four methods is respectively shaded in \textcolor{blue}{blue}, \textcolor{red}{red}, \textcolor{green}{green}, \textcolor{orange}{orange}; and a darker color indicates a higher Macro-F1.
}
\label{tab:imbalance}
\vspace{-0.1in}
\begin{tabular}{ll|ccc|ccc|ccc|ccc|cc}
\toprule
\multirow{2}{*}{Model} & \multirow{2}{*}{Loss Type} & \multicolumn{3}{c|}{Normal} & \multicolumn{3}{c|}{Mild} & \multicolumn{3}{c|}{Moderate} & \multicolumn{3}{c|}{Severe} & \multirow{2}{*}{Acc} & \multirow{2}{*}{Macro-F1} \\
& & Pre & Rec & F1 & Pre & Rec & F1 & Pre & Rec & F1 & Pre & Rec & F1 & \\
\midrule
\multirow{3}{*}{\makecell[l]{MLP\\(I-HOPE~\cite{roychowdhury2025predicting})}}
& Cross Entropy & 0.7223 & 0.8708 & 0.7896 & 0.5477 & 0.4655 & 0.5033 & 0.3750 & 0.1781 & 0.2415 & 0.9213 & 0.4916 & 0.6411 & 0.6662 & \cellcolor{blue!50}{0.5439} \\
& Weighted Loss & 0.8115 & 0.6933 & 0.7477 & 0.4945 & 0.4918 & 0.4931 & 0.2359 & 0.5068 & 0.3220 & 0.7750 & 0.5210 & 0.6231 & 0.6057 & \cellcolor{blue!80}0.5465 \\
& Focal Loss & 0.7723 & 0.7934 & 0.7827 & 0.5106 & 0.4832 & 0.4966 & 0.2430 & 0.2169 & 0.2292 & 0.4915 & 0.6092 & 0.5441 & 0.6378 & \cellcolor{blue!20}0.5132 \\
\hline
\multirow{3}{*}{TCN}
& Cross Entropy & 0.7142 & 0.8715 & 0.7850 & 0.5239 & 0.4015 & 0.4546 & 0.3597 & 0.2078 & 0.2634 & 0.6497 & 0.4832 & 0.5542 & 0.6484 & \cellcolor{red!60}0.5143 \\
& Weighted Loss & 0.8042 & 0.7003 & 0.7486 & 0.4996 & 0.4332 & 0.4641 & 0.2803 & 0.6119 & 0.3845 & 0.3909 & 0.5042 & 0.4404 & 0.5991 & \cellcolor{red!30}0.5094 \\
& Focal Loss & 0.7333 & 0.8396 & 0.7829 & 0.5344 & 0.4308 & 0.4770 & 0.3600 & 0.3288 & 0.3437 & 0.6722 & 0.5084 & 0.5789 & 0.6515 & \cellcolor{red!90}0.5456 \\
\hline
\multirow{3}{*}{\makecell[l]{LSTM\\+Attention}}
& Cross Entropy & 0.7617 & 0.7735 & 0.7675 & 0.5116 & 0.5674 & 0.5380 & 0.2878 & 0.2215 & 0.2503 & 0.7672 & 0.3739 & 0.5028 & 0.6430 & \cellcolor{green!80}0.5147 \\
& Weighted Loss & 0.9178 & 0.4923 & 0.6408 & 0.4311 & 0.7639 & 0.5512 & 0.3066 & 0.3493 & 0.3266 & 0.5169 & 0.5126 & 0.5148 & 0.5672 & \cellcolor{green!50}0.5084 \\
& Focal Loss & 0.7116 & 0.8295 & 0.7660 & 0.5032 & 0.4863 & 0.4946 & 0.3503 & 0.1373 & 0.1973 & 0.7541 & 0.4118 & 0.5327 & 0.6429 & \cellcolor{green!20}0.4977 \\
\hline
\multirow{3}{*}{Transformer}
& Cross Entropy & 0.7076 & 0.8880 & 0.7876 & 0.5404 & 0.4240 & 0.4752 & 0.3636 & 0.1461 & 0.2085 & 0.7760 & 0.4076 & 0.5344 & 0.6560 & \cellcolor{orange!50}0.5014 \\
& Weighted Loss & 0.8779 & 0.6569 & 0.7515 & 0.4895 & 0.6840 & 0.5706 & 0.3103 & 0.3493 & 0.3287 & 0.5777 & 0.6092 & 0.5930 & 0.6372 & \cellcolor{orange!70}{0.5610} \\
& Focal Loss & 0.8206 & 0.7335 & 0.7746 & 0.5130 & 0.5284 & 0.5206 & 0.2989 & 0.5479 & 0.3868 & 0.9213 & 0.4916 & 0.6411 & 0.6416 & \cellcolor{orange!90}{0.5808} \\
\bottomrule
\end{tabular}
\end{table}

\subsection{Discussion and Insights}

The experimental results across six research questions offer several key insights into the modeling of mental health forecasting using passive smartphone sensing data. These findings reveal the strengths and limitations of different modeling approaches (Traditional ML, DL, and LLMs) and provide guidance for future research in ubiquitous computing and digital mental health analytics.

(1) \textit{\textbf{DL Models as the Most Effective Forecasting Framework}}: Among all modeling approaches, DL models consistently achieve the highest overall performance in both Accuracy and Macro-F1, demonstrating their strong capacity to capture nonlinear temporal dependencies and complex behavioral dynamics. Transformers, in particular, stand out due to their ability to integrate multi-scale behavioral signals and retain long-term temporal information.
In contrast, traditional ML models such as SVM and Logistic Regression rely on handcrafted features and fixed temporal aggregation, which limits their ability to generalize under complex, temporally evolving conditions. However, their simplicity also makes them computationally efficient and interpretable, suggesting continued value in resource-constrained or real-time monitoring scenarios.

(2) \textit{\textbf{LLMs Show Promising Reasoning Ability but Limited Quantitative Accuracy}}:
Although LLMs demonstrate strong reasoning and generalization potential, their quantitative prediction accuracy remains below that of DL models and even most of ML models. This performance gap arises because current LLMs, even with PEFT and prompting, are primarily trained on textual corpora rather than multimodal or numeric data. As a result, they struggle to internalize fine-grained behavioral statistics that are essential for accurate forecasting.
Nevertheless, the results also show that LLMs capture semantic and contextual nuances more effectively than numerical aggregation methods. The superior performance of Similarity-based and Recency-based In-context Learning suggests that LLMs can reason about relative behavioral similarities when provided with well-structured examples, even if their absolute numeric estimation is weaker. This highlights their potential as behavioral reasoning engines rather than standalone predictors, complementing statistical or DL models in hybrid architectures.

(3) \textit{\textbf{Early Prediction Capability is Critical and Achievable}}: 
The incremental evaluation analysis reveals that DL models and LLMs possess stronger early prediction capabilities compared to traditional ML models. Their ability to achieve near-peak accuracy with only one week of data indicates that they effectively learn early behavioral precursors of mental health changes.
Interestingly, LLMs show performance fluctuations as the observation window expands, suggesting potential over-sensitivity to newly added, noisy data. This implies that shorter, well-curated behavioral contexts may be more informative than longer but noisier histories. The ability to make accurate early predictions is particularly valuable in mental health applications, as it supports timely interventions and risk mitigation before symptoms intensify.

(4) \textbf{\textit{Feature Representation (e.g., Granularity) are Crucial Design Choices}}:
Feature granularity significantly influences forecasting outcomes.
For traditional ML and MLP models, reduced (aggregated) feature dimensionality (5-Dimension) leads to better generalization, implying that high-dimensional sensing data often contain redundancy or noise. For sequential DL models, moderate dimension or temporal aggregation (5-Dimension or Weekly-level) improves robustness by smoothing out short-term behavioral fluctuations, while CNN-based temporal models (e.g., TCN) benefit from higher-frequency daily data due to their ability to capture local temporal patterns.
LLMs, however, perform best with simplified and semantically meaningful inputs, specifically the (5-Dimension, Weekly) setting, confirming that concise, aggregated textual representations are more compatible with their language reasoning mechanisms.

(5) \textit{\textbf{Importance of Personalized Modeling}}: Personalization proves to be a major factor in improving forecasting accuracy, especially for high-severity mental health conditions. Incorporating user identifiers or embeddings allows models to learn individual behavioral baselines and detect deviations that may indicate mental health deterioration.
DL models benefit most from personalization, achieving substantial Macro-F1 improvements (up to +0.3635), while traditional ML models show moderate but consistent gains. In contrast, LLMs do not benefit from simple personalization cues, such as user identifiers in prompts, due to their inability to retain long-term user histories or contextualized embeddings. This finding points toward the need for memory-augmented or history-aware LLMs that can integrate user-specific temporal trajectories in a principled manner.

(6) \textit{\textbf{Handling Class Imbalance Remains a Central Challenge}}: 
All modeling approaches show strong bias toward the Normal class, while performance for Mild, Moderate, and Severe states remains substantially lower. This confirms that imbalanced sample distributions and behavioral similarity between adjacent classes remain the primary bottlenecks.
Among the tested strategies, Focal Loss proves most effective for high-capacity models such as Transformers, as it dynamically emphasizes hard or minority samples without destabilizing optimization. In contrast, Weighted Cross-Entropy provides more stable but smaller improvements for simpler architectures like MLP. The findings underscore that model-specific loss adaptation is crucial to achieving balanced performance across mental health severity levels.

The findings collectively highlight several broader implications and future directions:
\begin{itemize}[leftmargin=*]
    \item Effective LLM Adaptation Strategies: The results demonstrate that In-Context Learning (ICL) consistently outperforms Parameter-Efficient Fine-Tuning (PEFT) methods (e.g., LoRA and Prompt Tuning) for this task. Future research should explore hybrid adaptation frameworks that jointly optimize ICL and PEFT to balance interpretability, computational efficiency, and task-specific alignment. Additionally, developing domain-aware prompts and behaviorally structured input representations may enable LLMs to reason more effectively over numerical and temporal data. Beyond these, knowledge distillation~\cite{du2025active} and advanced fine-tuning techniques like DPO~\cite{wei2025mirroring} and GRPO~\cite{du2025reinforcement} hold promise to improve domain adaptation while maintaining generalization.
    \item Hybrid Modeling: Combining DL models for quantitative forecasting with LLMs for semantic reasoning could yield synergistic effects, enabling interpretable and context-aware predictions.
    \item Data Efficiency and Explainability: Compact feature representations (e.g., weekly summaries or semantically compressed features) not only improve performance but also enhance model interpretability, an essential property in healthcare applications. 
    \item Personalized Mental Health AI: Future systems should integrate user-specific temporal embeddings, enabling adaptive modeling of individual behavioral trajectories.
    \item Toward Multimodal Integration: Extending current models to incorporate textual, physiological, and social context signals may further enhance generalization and enable more holistic mental health assessment~\cite{sahili2024multimodal,cabral20243m}.
\end{itemize}

In summary, the insights gained through our study advance the understanding of how traditional ML, DL and LLMs can be effectively leveraged for human-centric, interpretable, and proactive mental health forecasting using ubiquitous sensing data. 

\section{Conclusion}\label{sec:conclusion}

This study presents the first comprehensive comparison of traditional ML, DL, and LLM-based approaches for forecasting mental health from smartphone sensing data. Using the large-scale College Experience Sensing (CES) dataset, we systematically evaluate how different modeling strategies forecast mental health states using the phone sensing behavioral data. Our results show that DL approaches achieve the strongest overall performance, LLMs offer contextual reasoning advantages, and personalization markedly enhances forecasting accuracy, especially for severe mental health conditions. Together, these findings advance understanding of how computational models can forecast mental health changes before symptoms occur. More broadly, this work lays a foundation for adaptive, data-driven mental health technologies that promote early intervention, personalized support, and improved well-being, helping societies move toward more proactive and preventive mental health.


\bibliographystyle{ACM-Reference-Format}
\bibliography{ref,ref-forecast}

@inproceedings{umematsu2020forecasting,
  title={Forecasting stress, mood, and health from daytime physiology in office workers and students},
  author={Umematsu, Terumi and Sano, Akane and Taylor, Sara and Tsujikawa, Masanori and Picard, Rosalind W},
  booktitle={2020 42nd annual international conference of the IEEE Engineering in Medicine \& Biology Society (EMBC)},
  pages={5953--5957},
  year={2020},
  organization={IEEE}
}

@article{wei2025mirroring,
  title={Mirroring users: towards building preference-aligned user simulator with user feedback in recommendation},
  author={Wei, Tianjun and Guo, Huizhong and Du, Yingpeng and Sun, Zhu and Huang, Chen and Wang, Dongxia and Zhang, Jie},
  journal={arXiv preprint arXiv:2508.18142},
  year={2025}
}

@article{du2025reinforcement,
  title={Reinforcement speculative decoding for fast ranking},
  author={Du, Yingpeng and Wei, Tianjun and Sun, Zhu and Zhang, Jie},
  journal={arXiv preprint arXiv:2505.20316},
  year={2025}
}

@inproceedings{du2025active,
  title={Active large language model-based knowledge distillation for session-based recommendation},
  author={Du, Yingpeng and Sun, Zhu and Wang, Ziyan and Chua, Haoyan and Zhang, Jie and Ong, Yew-Soon},
  booktitle={Proceedings of the AAAI Conference on Artificial Intelligence},
  volume={39},
  number={11},
  pages={11607--11615},
  year={2025}
}

@article{langener2024predicting,
  title={Predicting mood based on the social context measured through the experience sampling method, digital phenotyping, and social networks},
  author={Langener, Anna M and Bringmann, Laura F and Kas, Martien J and Stulp, Gert},
  journal={Administration and Policy in Mental Health and Mental Health Services Research},
  volume={51},
  number={4},
  pages={455--475},
  year={2024},
  publisher={Springer}
}

@inproceedings{spathis2019sequence,
  title={Sequence multi-task learning to forecast mental wellbeing from sparse self-reported data},
  author={Spathis, Dimitris and Servia-Rodriguez, Sandra and Farrahi, Katayoun and Mascolo, Cecilia and Rentfrow, Jason},
  booktitle={Proceedings of the 25th ACM SIGKDD International Conference on Knowledge Discovery \& Data Mining},
  pages={2886--2894},
  year={2019}
}

@article{kathan2022personalised,
  title={Personalised depression forecasting using mobile sensor data and ecological momentary assessment},
  author={Kathan, Alexander and Harrer, Mathias and K{\"u}ster, Ludwig and Triantafyllopoulos, Andreas and He, Xiangheng and Milling, Manuel and Gerczuk, Maurice and Yan, Tianhao and Rajamani, Srividya Tirunellai and Heber, Elena and and Grossmann, Inga  and Ebert, David D.  and Schuller, Bj{\"o}rn W.},
  journal={Frontiers in digital health},
  volume={4},
  pages={964582},
  year={2022},
  publisher={Frontiers Media SA}
}

@article{paz2025emotion,
  title={Emotion forecasting: a transformer-based approach},
  author={Paz-Arbaizar, Leire and Lopez-Castroman, Jorge and Art{\'e}s-Rodr{\'\i}guez, Antonio and Olmos, Pablo M and Ram{\'\i}rez, David},
  journal={Journal of Medical Internet Research},
  volume={27},
  pages={e63962},
  year={2025},
  publisher={JMIR Publications Toronto, Canada}
}

@article{schat2024forecasting,
  title={Forecasting the onset of depression with limited baseline data only: a comparison of a person-specific and a multilevel modeling based exponentially weighted moving average approach.},
  author={Schat, Evelien and Tuerlinckx, Francis and Schreuder, Marieke J and De Ketelaere, Bart and Ceulemans, Eva},
  journal={Psychological Assessment},
  volume={36},
  number={6-7},
  pages={379},
  year={2024},
  publisher={American Psychological Association}
}

@inproceedings{roychowdhury2025predicting,
  author    = {Meghna Roy Chowdhury and Wei Xuan and Shreyas Sen and Yixue Zhao and Yi Ding},
  title     = {Predicting and understanding college student mental health with interpretable machine learning},
  booktitle = {2025 IEEE/ACM Conference on Connected Health: Applications, Systems and Engineering Technologies (CHASE)},
  year      = {2025},
}

@inproceedings{chen2016xgboost,
  title={Xgboost: a scalable tree boosting system},
  author={Chen, Tianqi and Guestrin, Carlos},
  booktitle={Proceedings of the 22nd acm sigkdd international conference on knowledge discovery and data mining},
  pages={785--794},
  year={2016}
}

@article{ke2017lightgbm,
  title={Lightgbm: a highly efficient gradient boosting decision tree},
  author={Ke, Guolin and Meng, Qi and Finley, Thomas and Wang, Taifeng and Chen, Wei and Ma, Weidong and Ye, Qiwei and Liu, Tie-Yan},
  journal={Advances in neural information processing systems},
  volume={30},
  year={2017}
}

@book{hosmer2013applied,
  title={Applied logistic regression},
  author={Hosmer Jr, David W and Lemeshow, Stanley and Sturdivant, Rodney X},
  year={2013},
  publisher={John Wiley \& Sons}
}

@article{brereton2010support,
  title={Support vector machines for classification and regression},
  author={Brereton, Richard G and Lloyd, Gavin R},
  journal={Analyst},
  volume={135},
  number={2},
  pages={230--267},
  year={2010},
  publisher={Royal Society of Chemistry}
}

@article{song2015decision,
  title={Decision tree methods: applications for classification and prediction},
  author={Song, Yan-Yan and Lu, Ying},
  journal={Shanghai archives of psychiatry},
  year={2015},
  publisher={Shanghai Municipal Bureau of Publishing}
}

@article{breiman2001random,
  title={Random forests},
  author={Breiman, Leo},
  journal={Machine learning},
  volume={45},
  number={1},
  pages={5--32},
  year={2001},
  publisher={Springer}
}

@article{scikit-learn,
  title={Scikit-learn: machine learning in python},
  author={Pedregosa, Fabian and Varoquaux, Ga{\"e}l and Gramfort, Alexandre and Michel, Vincent and Thirion, Bertrand and Grisel, Olivier and Blondel, Mathieu and Prettenhofer, Peter and Weiss, Ron and Dubourg, Vincent and Vanderplas, Jake and Passos, Alexandre and Cournapeau, David and Brucher, Matthieu and Perrot, Matthieu and Duchesnay, {\'E}douard},
  journal={Journal of Machine Learning Research},
  volume={12},
  pages={2825--2830},
  year={2011}
}

@inproceedings{paszke2019pytorch,
  title={{PyTorch}: an imperative style, high-performance deep learning library},
  author={Paszke, Adam and Gross, Sam and Massa, Francisco and Lerer, Adam and Bradbury, James and Chanan, Gregory and Killeen, Trevor and Lin, Zeming and Gimelshein, Natalia and Antiga, Luca and Desmaison, Alban and Kopf, Andreas and Yang, Edward and DeVito, Zachary and Raison, Martin and Tejani, Alykhan and Chilamkurthy, Sasank and Steiner, Benoit and Fang, Lu and Bai, Junjie and Chintala, Soumith},
  booktitle={Advances in Neural Information Processing Systems 32},
  editor={Wallach, H. and Larochelle, H. and Beygelzimer, A. and d'Alch{\'e}-Buc, F. and Fox, E. and Garnett, R.},
  pages={8024--8035},
  year={2019},
}

@article{qwen2023technical,
  title   = {Qwen technical report},
  author  = {Qwen Team},
  journal = {arXiv preprint arXiv:2309.16609},
  year    = {2023},
}

@misc{openai_gpt4.1_2025,
  title        = {{GPT-4.1} Technical report},
  author       = {{OpenAI}},
  howpublished = {\url{https://openai.com/index/gpt-4-1/}},
  year         = {2025},
}

@inproceedings{wolf-etal-2020-transformers,
  title = {HuggingFace's Transformers: state-of-the-art natural language processing},
  author = {Wolf, Thomas and Debut, Lysandre and Sanh, Victor and Chaumond, Julien and Delangue, Clement and Moi, Anthony and Cistac, Pierric and Rault, Tim and Louf, Rémi and Funtowicz, Morgan and Davison, Joe and Shleifer, Sam and von Platen, Patrick and Ma, Clara and Jernite, Yacine and Plu, Julien and Xu, Canwen and Le Scao, Teven and Gugger, Sylvain and Drame, Mariama and Lhoest, Quentin and Rush, Alexander M.},
  booktitle = {Proceedings of the 2020 Conference on Empirical Methods in Natural Language Processing: System Demonstrations},
  year = {2020},
  publisher = {Association for Computational Linguistics},
  pages = {38--45},
}

@article{nelson2019accuracy,
  title={Accuracy of consumer wearable heart rate measurement during an ecologically valid 24-hour period: intraindividual validation study},
  author={Nelson, Benjamin W and Allen, Nicholas B},
  journal={JMIR mHealth and uHealth},
  volume={7},
  number={3},
  pages={e10828},
  year={2019},
  publisher={JMIR Publications Toronto, Canada}
}

@article{bardram2022sensing,
  title={From sensing to acting—can pervasive computing change the world?},
  author={Bardram, Jakob E},
  journal={IEEE Pervasive Computing},
  volume={21},
  number={3},
  pages={17--23},
  year={2022},
  publisher={IEEE}
}

@article{boe2019automating,
  title={Automating sleep stage classification using wireless, wearable sensors},
  author={Boe, Alexander J and McGee Koch, Lori L and O’Brien, Megan K and Shawen, Nicholas and Rogers, John A and Lieber, Richard L and Reid, Kathryn J and Zee, Phyllis C and Jayaraman, Arun},
  journal={NPJ digital medicine},
  volume={2},
  number={1},
  pages={131},
  year={2019},
  publisher={Nature Publishing Group UK London}
}

@inproceedings{abdullah2014towards,
  title={Towards circadian computing: "early to bed and early to rise" makes some of us unhealthy and sleep deprived},
  author={Abdullah, Saeed and Matthews, Mark and Murnane, Elizabeth L and Gay, Geri and Choudhury, Tanzeem},
  booktitle={Proceedings of the 2014 ACM international joint conference on pervasive and ubiquitous computing},
  pages={673--684},
  year={2014}
}

@article{coravos2019developing,
  title={Developing and adopting safe and effective digital biomarkers to improve patient outcomes},
  author={Coravos, Andrea and Khozin, Sean and Mandl, Kenneth D},
  journal={NPJ digital medicine},
  volume={2},
  number={1},
  pages={14},
  year={2019},
  publisher={Nature Publishing Group UK London}
}

@article{dunn2021wearable,
  title={Wearable sensors enable personalized predictions of clinical laboratory measurements},
  author={Dunn, Jessilyn and Kidzinski, Lukasz and Runge, Ryan and Witt, Daniel and Hicks, Jennifer L and Sch{\"u}ssler-Fiorenza Rose, Sophia Miryam and Li, Xiao and Bahmani, Amir and Delp, Scott L and Hastie, Trevor and Snyder, Michael P.},
  journal={Nature medicine},
  volume={27},
  number={6},
  pages={1105--1112},
  year={2021},
  publisher={Nature Publishing Group US New York}
}

@article{master2022association,
  title={Association of step counts over time with the risk of chronic disease in the All of Us Research Program},
  author={Master, Hiral and Annis, Jeffrey and Huang, Shi and Beckman, Joshua A and Ratsimbazafy, Francis and Marginean, Kayla and Carroll, Robert and Natarajan, Karthik and Harrell, Frank E and Roden, Dan M and Harris, Paul and Brittain, Evan L.},
  journal={Nature medicine},
  volume={28},
  number={11},
  pages={2301--2308},
  year={2022},
  publisher={Nature Publishing Group US New York}
}

@article{nepal2024social,
  title={Social isolation and serious mental illness: the role of context-aware mobile interventions},
  author={Nepal, Subigya and Pillai, Arvind and Parrish, Emma M and Holden, Jason and Depp, Colin and Campbell, Andrew T and Granholm, Eric L},
  journal={IEEE pervasive computing},
  volume={23},
  number={1},
  pages={46--56},
  year={2024},
  publisher={IEEE}
}

@article{xu2022globem,
  title={GLOBEM dataset: multi-year datasets for longitudinal human behavior modeling generalization},
  author={Xu, Xuhai and Zhang, Han and Sefidgar, Yasaman and Ren, Yiyi and Liu, Xin and Seo, Woosuk and Brown, Jennifer and Kuehn, Kevin and Merrill, Mike and Nurius, Paula and others},
  journal={Advances in neural information processing systems},
  volume={35},
  pages={24655--24692},
  year={2022}
}

@article{xu2023globem,
  title={Globem: cross-dataset generalization of longitudinal human behavior modeling},
  author={Xu, Xuhai and Liu, Xin and Zhang, Han and Wang, Weichen and Nepal, Subigya and Sefidgar, Yasaman and Seo, Woosuk and Kuehn, Kevin S and Huckins, Jeremy F and Morris, Margaret E and others},
  journal={Proceedings of the ACM on Interactive, Mobile, Wearable and Ubiquitous Technologies},
  volume={6},
  number={4},
  pages={1--34},
  year={2023},
  publisher={ACM New York, NY, USA}
}

@article{nahum2017just,
  title={Just-in-time adaptive interventions (JITAIs) in mobile health: key components and design principles for ongoing health behavior support},
  author={Nahum-Shani, Inbal and Smith, Shawna N and Spring, Bonnie J and Collins, Linda M and Witkiewitz, Katie and Tewari, Ambuj and Murphy, Susan A},
  journal={Annals of behavioral medicine},
  year={2017},
  publisher={Springer}
}

@article{klasnja2019efficacy,
  title={Efficacy of contextually tailored suggestions for physical activity: a micro-randomized optimization trial of HeartSteps},
  author={Klasnja, Predrag and Smith, Shawna and Seewald, Nicholas J and Lee, Andy and Hall, Kelly and Luers, Brook and Hekler, Eric B and Murphy, Susan A},
  journal={Annals of Behavioral Medicine},
  volume={53},
  number={6},
  pages={573--582},
  year={2019},
  publisher={Oxford University Press US}
}

@article{stamatis2023specific,
  title={Specific associations of passively sensed smartphone data with future symptoms of avoidance, fear, and physiological distress in social anxiety},
  author={Stamatis, Caitlin A and Liu, Tingting and Meyerhoff, Jonah and Meng, Yixuan and Cho, Young Min and Karr, Chris J and Curtis, Brenda L and Ungar, Lyle H and Mohr, David C},
  journal={Internet Interventions},
  volume={34},
  pages={100683},
  year={2023},
  publisher={Elsevier}
}

@incollection{vaid2021ubiquitous,
  title={Ubiquitous computing for person-environment research: opportunities, considerations, and future directions},
  author={Vaid, Sumer S and Abdullah, Saeed and Thomaz, Edison and Harari, Gabriella M},
  booktitle={Measuring and modeling persons and situations},
  pages={103--143},
  year={2021},
  publisher={Elsevier}
}

@article{mohd2024relationship,
  title={Relationship of screen time with anxiety, depression, and sleep quality among adolescents: a cross-sectional study},
  author={Mohd Saat, Nur Zakiah and Hanawi, Siti Aishah and Hanafiah, Hazlenah and Ahmad, Mahadir and Farah, Nor MF and Abdul Rahman, Nur Ain Atikah},
  journal={Frontiers in Public Health},
  volume={12},
  pages={1459952},
  year={2024},
  publisher={Frontiers Media SA}
}

@article{zhang2024reproducible,
  title={A reproducible stress prediction pipeline with mobile sensor data},
  author={Zhang, Panyu and Jung, Gyuwon and Alikhanov, Jumabek and Ahmed, Uzair and Lee, Uichin},
  journal={Proceedings of the ACM on interactive, mobile, wearable and ubiquitous technologies},
  volume={8},
  number={3},
  pages={1--35},
  year={2024},
  publisher={ACM New York, NY, USA}
}

@article{nepal2024capturing,
  title={Capturing the college experience: a four-year mobile sensing study of mental health, resilience and behavior of college students during the pandemic},
  author={Nepal, Subigya and Liu, Wenjun and Pillai, Arvind and Wang, Weichen and Vojdanovski, Vlado and Huckins, Jeremy F and Rogers, Courtney and Meyer, Meghan L and Campbell, Andrew T},
  journal={Proceedings of the ACM on interactive, mobile, wearable and ubiquitous technologies},
  volume={8},
  number={1},
  pages={1--37},
  year={2024},
  publisher={ACM New York, NY, USA}
}

@inproceedings{wang2014studentlife,
  title={StudentLife: assessing mental health, academic performance and behavioral trends of college students using smartphones},
  author={Wang, Rui and Chen, Fanglin and Chen, Zhenyu and Li, Tianxing and Harari, Gabriella and Tignor, Stefanie and Zhou, Xia and Ben-Zeev, Dror and Campbell, Andrew T},
  booktitle={Proceedings of the 2014 ACM international joint conference on pervasive and ubiquitous computing},
  pages={3--14},
  year={2014}
}

@article{adler2021identifying,
  title={Identifying mobile sensing indicators of stress-resilience},
  author={Adler, Daniel A and Tseng, Vincent W-S and Qi, Gengmo and Scarpa, Joseph and Sen, Srijan and Choudhury, Tanzeem},
  journal={Proceedings of the ACM on interactive, mobile, wearable and ubiquitous technologies},
  volume={5},
  number={2},
  pages={1--32},
  year={2021},
  publisher={ACM New York, NY, USA}
}

@article{barnett2018relapse,
  title={Relapse prediction in schizophrenia through digital phenotyping: a pilot study},
  author={Barnett, Ian and Torous, John and Staples, Patrick and Sandoval, Luis and Keshavan, Matcheri and Onnela, Jukka-Pekka},
  journal={Neuropsychopharmacology},
  volume={43},
  number={8},
  pages={1660--1666},
  year={2018},
  publisher={Springer International Publishing Cham}
}

@inproceedings{wang2016crosscheck,
  title={CrossCheck: toward passive sensing and detection of mental health changes in people with schizophrenia},
  author={Wang, Rui and Aung, Min SH and Abdullah, Saeed and Brian, Rachel and Campbell, Andrew T and Choudhury, Tanzeem and Hauser, Marta and Kane, John and Merrill, Michael and Scherer, Emily A and  Tseng, Vincent W. S.and Ben-Zeev, Dror},
  booktitle={Proceedings of the 2016 ACM international joint conference on pervasive and ubiquitous computing},
  pages={886--897},
  year={2016}
}

@article{adler2024measuring,
  title={Measuring algorithmic bias to analyze the reliability of AI tools that predict depression risk using smartphone sensed-behavioral data},
  author={Adler, Daniel A and Stamatis, Caitlin A and Meyerhoff, Jonah and Mohr, David C and Wang, Fei and Aranovich, Gabriel J and Sen, Srijan and Choudhury, Tanzeem},
  journal={npj Mental Health Research},
  volume={3},
  number={1},
  pages={17},
  year={2024},
  publisher={Nature Publishing Group UK London}
}

@article{meegahapola2023generalization,
  title={Generalization and personalization of mobile sensing-based mood inference models: an analysis of college students in eight countries},
  author={Meegahapola, Lakmal and Droz, William and Kun, Peter and de G\"{o}tzen, Amalia and Nutakki, Chaitanya and Diwakar, Shyam and Correa, Salvador Ruiz and Song, Donglei and Xu, Hao and Bidoglia, Miriam and Gaskell, George and Chagnaa, Altangerel and Ganbold, Amarsanaa and Zundui, Tsolmon and Caprini, Carlo and Miorandi, Daniele and Hume, Alethia and Zarza, Jose Luis and Cernuzzi, Luca and Bison, Ivano and Britez, Marcelo Rodas and Busso, Matteo and Chenu-Abente, Ronald and G\"{u}nel, Can and Giunchiglia, Fausto and Schelenz, Laura and Gatica-Perez, Daniel},
  journal={Proceedings of the ACM on interactive, mobile, wearable and ubiquitous technologies},
  volume={6},
  number={4},
  pages={1--32},
  year={2023},
  publisher={ACM New York, NY, USA}
}

@article{meegahapola2024m3bat,
  title={M3BAT: unsupervised domain adaptation for multimodal mobile sensing with multi-branch adversarial training},
  author={Meegahapola, Lakmal and Hassoune, Hamza and Gatica-Perez, Daniel},
  journal={Proceedings of the ACM on Interactive, Mobile, Wearable and Ubiquitous Technologies},
  volume={8},
  number={2},
  pages={1--30},
  year={2024},
  publisher={ACM New York, NY, USA}
}

@article{kocon2023chatgpt,
  title={ChatGPT: jack of all trades, master of none},
  author = {Koco\'{n}, Jan and Cichecki, Igor and Kaszyca, Oliwier and Kochanek, Mateusz and Szyd\l{}o, Dominika and Baran, Joanna and Bielaniewicz, Julita and Gruza, Marcin and Janz, Arkadiusz and Kanclerz, Kamil and Koco\'{n}, Anna and Koptyra, Bart\l{}omiej and Mieleszczenko-Kowszewicz, Wiktoria and Mi\l{}kowski, Piotr and Oleksy, Marcin and Piasecki, Maciej and Radli\'{n}ski, \L{}ukasz and Wojtasik, Konrad and Wo\'{z}niak, Stanis\l{}aw and Kazienko, Przemys\l{}aw},
  journal={Information fusion},
  volume={99},
  pages={101861},
  year={2023},
  publisher={Elsevier}
}

@inproceedings{qin2023chatgpt,
    title = "Is ChatGPT a general-purpose natural language processing task solver?",
    author = "Qin, Chengwei  and
      Zhang, Aston  and
      Zhang, Zhuosheng  and
      Chen, Jiaao  and
      Yasunaga, Michihiro  and
      Yang, Diyi",
    editor = "Bouamor, Houda  and
      Pino, Juan  and
      Bali, Kalika",
    booktitle = "Proceedings of the 2023 Conference on Empirical Methods in Natural Language Processing",
    month = dec,
    year = "2023",
    address = "Singapore",
    publisher = "Association for Computational Linguistics",
    url = "https://aclanthology.org/2023.emnlp-main.85/",
    doi = "10.18653/v1/2023.emnlp-main.85",
    pages = "1339--1384",
}

@article{zhong2023can,
  title={Can chatgpt understand too? a comparative study on chatgpt and fine-tuned bert},
  author={Zhong, Qihuang and Ding, Liang and Liu, Juhua and Du, Bo and Tao, Dacheng},
  journal={arXiv preprint arXiv:2302.10198},
  year={2023}
}

@article{lamichhane2023evaluation,
  title={Evaluation of chatgpt for NLP-based mental health applications},
  author={Lamichhane, Bishal},
  journal={arXiv preprint arXiv:2303.15727},
  year={2023}
}

@article{amin2023affective,
author={Amin, Mostafa M. and Cambria, Erik and Schuller, Bjorn W.},
journal={IEEE Intelligent Systems },
title={Will affective computing emerge from foundation models and general artificial intelligence? A first evaluation of ChatGPT},
year={2023},
volume={38},
number={02},
ISSN={1941-1294},
pages={15-23},
publisher={IEEE Computer Society},
}

@article{yang2023evaluations,
  title={On the evaluations of chatgpt and emotion-enhanced prompting for mental health analysis},
  author={Yang, Kailai and Ji, Shaoxiong and Zhang, Tianlin and Xie, Qianqian and Ananiadou, Sophia},
  journal={arXiv preprint arXiv:2304.03347},
  volume={4},
  year={2023}
}

@inproceedings{yang2023towards,
    title = "Towards interpretable mental health analysis with large language models",
    author = "Yang, Kailai  and
      Ji, Shaoxiong  and
      Zhang, Tianlin  and
      Xie, Qianqian  and
      Kuang, Ziyan  and
      Ananiadou, Sophia",
    editor = "Bouamor, Houda  and
      Pino, Juan  and
      Bali, Kalika",
    booktitle = "Proceedings of the 2023 Conference on Empirical Methods in Natural Language Processing",
    month = dec,
    year = "2023",
    address = "Singapore",
    publisher = "Association for Computational Linguistics",
}

@inproceedings{yang2024mentallama,
  title={MentaLLaMA: interpretable mental health analysis on social media with large language models},
  author={Yang, Kailai and Zhang, Tianlin and Kuang, Ziyan and Xie, Qianqian and Huang, Jimin and Ananiadou, Sophia},
  booktitle={Proceedings of the ACM Web Conference 2024},
  pages={4489--4500},
  year={2024}
}

@article{xu2024mental,
  title={Mental-llm: leveraging large language models for mental health prediction via online text data},
  author={Xu, Xuhai and Yao, Bingsheng and Dong, Yuanzhe and Gabriel, Saadia and Yu, Hong and Hendler, James and Ghassemi, Marzyeh and Dey, Anind K and Wang, Dakuo},
  journal={Proceedings of the ACM on Interactive, Mobile, Wearable and Ubiquitous Technologies},
  volume={8},
  number={1},
  pages={1--32},
  year={2024},
  publisher={ACM New York, NY, USA}
}

@InProceedings{kim2024healthllm,
  title = 	 {Health-LLM: large language models for health prediction via wearable sensor data},
  author =       {Kim, Yubin and Xu, Xuhai and McDuff, Daniel and Breazeal, Cynthia and Park, Hae Won},
  booktitle = 	 {Proceedings of the fifth Conference on Health, Inference, and Learning},
  pages = 	 {522--539},
  year = 	 {2024},
  editor = 	 {Pollard, Tom and Choi, Edward and Singhal, Pankhuri and Hughes, Michael and Sizikova, Elena and Mortazavi, Bobak and Chen, Irene and Wang, Fei and Sarker, Tasmie and McDermott, Matthew and Ghassemi, Marzyeh},
  volume = 	 {248},
  series = 	 {Proceedings of Machine Learning Research},
  month = 	 {27--28 Jun},
  publisher =    {PMLR},
}

@article{huckins2020mental,
  title={Mental health and behavior of college students during the early phases of the COVID-19 pandemic: longitudinal smartphone and ecological momentary assessment study},
  author={Huckins, Jeremy F and DaSilva, Alex W and Wang, Weichen and Hedlund, Elin and Rogers, Courtney and Nepal, Subigya K and Wu, Jialing and Obuchi, Mikio and Murphy, Eilis I and Meyer, Meghan L and Wagner, Dylan D and Holtzheimer, Paul E and Campbell, Andrew T},
  journal={Journal of medical Internet research},
  volume={22},
  number={6},
  pages={e20185},
  year={2020},
  publisher={JMIR Publications Toronto, Canada}
}

@inproceedings{nepal2022covid,
  title={COVID student study: a year in the life of college students during the COVID-19 pandemic through the lens of mobile phone sensing},
  author={Nepal, Subigya and Wang, Weichen and Vojdanovski, Vlado and Huckins, Jeremy F and Dasilva, Alex and Meyer, Meghan and Campbell, Andrew},
  booktitle={Proceedings of the 2022 CHI conference on human factors in computing systems},
  pages={1--19},
  year={2022}
}

@article{aung2016leveraging,
  title={Leveraging multi-modal sensing for mobile health: a case review in chronic pain},
  author={Aung, Min S Hane and Alquaddoomi, Faisal and Hsieh, Cheng-Kang and Rabbi, Mashfiqui and Yang, Longqi and Pollak, John P and Estrin, Deborah and Choudhury, Tanzeem},
  journal={IEEE journal of selected topics in signal processing},
  volume={10},
  number={5},
  pages={962--974},
  year={2016},
  publisher={IEEE}
}

@inproceedings{canzian2015trajectories,
  title={Trajectories of depression: unobtrusive monitoring of depressive states by means of smartphone mobility traces analysis},
  author={Canzian, Luca and Musolesi, Mirco},
  booktitle={Proceedings of the 2015 ACM international joint conference on pervasive and ubiquitous computing},
  pages={1293--1304},
  year={2015}
}

@article{ben2015next,
  title={Next-generation psychiatric assessment: using smartphone sensors to monitor behavior and mental health.},
  author={Ben-Zeev, Dror and Scherer, Emily A and Wang, Rui and Xie, Haiyi and Campbell, Andrew T},
  journal={Psychiatric rehabilitation journal},
  volume={38},
  number={3},
  pages={218},
  year={2015},
  publisher={Educational Publishing Foundation}
}

@article{saeb2015mobile,
  title={Mobile phone sensor correlates of depressive symptom severity in daily-life behavior: an exploratory study},
  author={Saeb, Sohrab and Zhang, Mi and Karr, Christopher J and Schueller, Stephen M and Corden, Marya E and Kording, Konrad P and Mohr, David C},
  journal={Journal of medical Internet research},
  volume={17},
  number={7},
  pages={e4273},
  year={2015},
  publisher={JMIR Publications Inc., Toronto, Canada}
}

@article{chikersal2021detecting,
  title={Detecting depression and predicting its onset using longitudinal symptoms captured by passive sensing: a machine learning approach with robust feature selection},
  author={Chikersal, Prerna and Doryab, Afsaneh and Tumminia, Michael and Villalba, Daniella K and Dutcher, Janine M and Liu, Xinwen and Cohen, Sheldon and Creswell, Kasey G and Mankoff, Jennifer and Creswell, J David and others},
  journal={ACM Transactions on Computer-Human Interaction (TOCHI)},
  volume={28},
  number={1},
  pages={1--41},
  year={2021},
  publisher={ACM New York, NY, USA}
}

@inproceedings{farhan2016behavior,
  title={Behavior vs. introspection: refining prediction of clinical depression via smartphone sensing data},
  author={Farhan, Asma Ahmad and Yue, Chaoqun and Morillo, Reynaldo and Ware, Shweta and Lu, Jin and Bi, Jinbo and Kamath, Jayesh and Russell, Alexander and Bamis, Athanasios and Wang, Bing},
  booktitle={2016 IEEE wireless health (WH)},
  pages={1--8},
  year={2016},
  organization={IEEE}
}

@article{wang2018tracking,
  title={Tracking depression dynamics in college students using mobile phone and wearable sensing},
  author={Wang, Rui and Wang, Weichen and DaSilva, Alex and Huckins, Jeremy F and Kelley, William M and Heatherton, Todd F and Campbell, Andrew T},
  journal={Proceedings of the ACM on Interactive, Mobile, Wearable and Ubiquitous Technologies},
  volume={2},
  number={1},
  pages={1--26},
  year={2018},
  publisher={ACM New York, NY, USA}
}

@article{xu2019leveraging,
  title={Leveraging routine behavior and contextually-filtered features for depression detection among college students},
  author={Xu, Xuhai and Chikersal, Prerna and Doryab, Afsaneh and Villalba, Daniella K and Dutcher, Janine M and Tumminia, Michael J and Althoff, Tim and Cohen, Sheldon and Creswell, Kasey G and Creswell, J David and Mankof, Jennifer and Dey, Anind K.},
  journal={Proceedings of the ACM on Interactive, Mobile, Wearable and Ubiquitous Technologies},
  volume={3},
  number={3},
  pages={1--33},
  year={2019},
  publisher={ACM New York, NY, USA}
}

@article{fukazawa2020smartphone,
  title={Smartphone-based mental state estimation: a survey from a machine learning perspective},
  author={Fukazawa, Yusuke and Yamamoto, Naoki and Hamatani, Takashi and Ochiai, Keiichi and Uchiyama, Akira and Ohta, Ken},
  journal={Journal of Information Processing},
  volume={28},
  pages={16--30},
  year={2020},
  publisher={Information Processing Society of Japan}
}

@article{vos2023generalizable,
  title={Generalizable machine learning for stress monitoring from wearable devices: a systematic literature review},
  author={Vos, Gideon and Trinh, Kelly and Sarnyai, Zoltan and Azghadi, Mostafa Rahimi},
  journal={International Journal of Medical Informatics},
  volume={173},
  pages={105026},
  year={2023},
  publisher={Elsevier}
}

@article{yang2022survey,
  title={Survey on emotion sensing using mobile devices},
  author={Yang, Kangning and Tag, Benjamin and Wang, Chaofan and Gu, Yue and Sarsenbayeva, Zhanna and Dingler, Tilman and Wadley, Greg and Goncalves, Jorge},
  journal={IEEE Transactions on Affective Computing},
  volume={14},
  number={4},
  pages={2678--2696},
  year={2022},
  publisher={IEEE}
}

@article{khwaja2019modeling,
  title={Modeling personality vs. modeling personalidad: In-the-wild mobile data analysis in five countries suggests cultural impact on personality models},
  author={Khwaja, Mohammed and Vaid, Sumer S and Zannone, Sara and Harari, Gabriella M and Faisal, A Aldo and Matic, Aleksandar},
  journal={Proceedings of the ACM on Interactive, Mobile, Wearable and Ubiquitous Technologies},
  volume={3},
  number={3},
  pages={1--24},
  year={2019},
  publisher={ACM New York, NY, USA}
}

@article{adler2022machine,
  title={Machine learning for passive mental health symptom prediction: generalization across different longitudinal mobile sensing studies},
  author={Adler, Daniel A and Wang, Fei and Mohr, David C and Choudhury, Tanzeem},
  journal={Plos one},
  volume={17},
  number={4},
  pages={e0266516},
  year={2022},
  publisher={Public Library of Science San Francisco, CA USA}
}

@article{kroenke2009ultra,
  title={An ultra-brief screening scale for anxiety and depression: the PHQ--4},
  author={Kroenke, Kurt and Spitzer, Robert L and Williams, Janet BW and L{\"o}we, Bernd},
  journal={Psychosomatics},
  volume={50},
  number={6},
  pages={613--621},
  year={2009},
  publisher={Elsevier}
}

@article{aurelio2019learning,
  title={Learning from imbalanced data sets with weighted cross-entropy function},
  author={Aurelio, Yuri Sousa and De Almeida, Gustavo Matheus and de Castro, Cristiano Leite and Braga, Antonio Padua},
  journal={Neural processing letters},
  volume={50},
  number={2},
  pages={1937--1949},
  year={2019},
  publisher={Springer}
}

@InProceedings{lin2017focal,
author = {Lin, Tsung-Yi and Goyal, Priya and Girshick, Ross and He, Kaiming and Dollar, Piotr},
title = {Focal loss for dense object detection},
booktitle = {Proceedings of the IEEE International Conference on Computer Vision (ICCV)},
month = {Oct},
year = {2017}
}

@article{zhu2023stress,
  title={Stress detection through wrist-based electrodermal activity monitoring and machine learning},
  author={Zhu, Lili and Spachos, Petros and Ng, Pai Chet and Yu, Yuanhao and Wang, Yang and Plataniotis, Konstantinos and Hatzinakos, Dimitrios},
  journal={IEEE Journal of Biomedical and Health Informatics},
  volume={27},
  number={5},
  pages={2155--2165},
  year={2023},
  publisher={IEEE}
}

@article{chen2025unveiling,
  title={Unveiling the landscape of clinical depression assessment: from behavioral signatures to psychiatric reasoning},
  author={Chen, Zhuang and Bi, Guanqun and Zhang, Wen and Hu, Jiawei and Wang, Aoyun and Xiao, Xiyao and Feng, Kun and Huang, Minlie},
  journal={arXiv preprint arXiv:2508.04531},
  year={2025}
}

@article{dettmers2023qlora,
  title={Qlora: efficient finetuning of quantized llms},
  author={Dettmers, Tim and Pagnoni, Artidoro and Holtzman, Ari and Zettlemoyer, Luke},
  journal={Advances in neural information processing systems},
  volume={36},
  pages={10088--10115},
  year={2023}
}

@inproceedings{liliang2021prefix,
    title = "Prefix-Tuning: optimizing continuous prompts for generation",
    author = "Li, Xiang Lisa  and
      Liang, Percy",
    editor = "Zong, Chengqing  and
      Xia, Fei  and
      Li, Wenjie  and
      Navigli, Roberto",
    booktitle = "Proceedings of the 59th Annual Meeting of the Association for Computational Linguistics and the 11th International Joint Conference on Natural Language Processing (Volume 1: Long Papers)",
    year = "2021",
    publisher = "Association for Computational Linguistics",
    pages = "4582--4597",
}

@article{han2024parameter,
  title={Parameter-efficient fine-tuning for large models: a comprehensive survey},
  author={Han, Zeyu and Gao, Chao and Liu, Jinyang and Zhang, Jeff and Zhang, Sai Qian},
  journal={arXiv preprint arXiv:2403.14608},
  year={2024}
}

@article{vaswani2017attention,
  title={Attention is all you need},
  author={Vaswani, Ashish and Shazeer, Noam and Parmar, Niki and Uszkoreit, Jakob and Jones, Llion and Gomez, Aidan N and Kaiser, {\L}ukasz and Polosukhin, Illia},
  journal={Advances in neural information processing systems},
  volume={30},
  year={2017}
}

@article{yu2019review,
  title={A review of recurrent neural networks: LSTM cells and network architectures},
  author={Yu, Yong and Si, Xiaosheng and Hu, Changhua and Zhang, Jianxun},
  journal={Neural computation},
  volume={31},
  number={7},
  pages={1235--1270},
  year={2019}
}

@article{bai2018empirical,
  title={An empirical evaluation of generic convolutional and recurrent networks for sequence modeling},
  author={Bai, Shaojie and Kolter, J Zico and Koltun, Vladlen},
  journal={arXiv preprint arXiv:1803.01271},
  year={2018}
}

@inproceedings{umematsu2019daytime,
  title={Daytime data and LSTM can forecast tomorrow’s stress, health, and happiness},
  author={Umematsu, Terumi and Sano, Akane and Picard, Rosalind W},
  booktitle={2019 41st Annual International Conference of the IEEE Engineering in Medicine and Biology Society (EMBC)},
  pages={2186--2190},
  year={2019},
  organization={IEEE}
}

@article{lewis2023mixed,
  title={Mixed effects random forests for personalised predictions of clinical depression severity},
  author={Lewis, Robert A and Ghandeharioun, Asma and Fedor, Szymon and Pedrelli, Paola and Picard, Rosalind and Mischoulon, David},
  journal={arXiv preprint arXiv:2301.09815},
  year={2023}
}

@article{shen2025passive,
  title={Passive sensing for mental health monitoring using machine learning with wearables and smartphones: scoping review},
  author={Shen, ShiYing and Qi, Wenhao and Zeng, Jianwen and Li, Sixie and Liu, Xin and Zhu, Xiaohong and Dong, Chaoqun and Wang, Bin and Shi, Yankai and Yao, Jiani and others},
  journal={Journal of Medical Internet Research},
  volume={27},
  pages={e77066},
  year={2025},
  publisher={JMIR Publications Toronto, Canada}
}

@article{opoku2021predicting,
  title={Predicting depression from smartphone behavioral markers using machine learning methods, hyperparameter optimization, and feature importance analysis: exploratory study},
  author={Opoku Asare, Kennedy and Terhorst, Yannik and Vega, Julio and Peltonen, Ella and Lagerspetz, Eemil and Ferreira, Denzil},
  journal={JMIR mHealth and uHealth},
  volume={9},
  number={7},
  pages={e26540},
  year={2021},
  publisher={JMIR Publications Toronto, Canada}
}

@article{sahili2024multimodal,
  title={Multimodal machine learning in mental health: a survey of data, algorithms, and challenges},
  author={Sahili, Zahraa Al and Patras, Ioannis and Purver, Matthew},
  journal={arXiv preprint arXiv:2407.16804},
  year={2024}
}

@inproceedings{cabral20243m,
  title={3M-Health: multimodal multi-teacher knowledge distillation for mental health detection},
  author={Cabral, Rina Carines and Luo, Siwen and Poon, Josiah and Han, Soyeon Caren},
  booktitle={Proceedings of the 33rd ACM International Conference on Information and Knowledge Management},
  pages={152--162},
  year={2024}
}

\end{document}